\documentclass[10pt,twocolumn,letterpaper]{article}

\usepackage[accsupp]{axessibility}
\usepackage[pagenumbers]{iccv} 

\usepackage{multirow}
\usepackage{physics}

\newcommand{\resultcellThreeCol}[1]{
\begin{tabular}{@{}c@{}}
    \includegraphics[width=0.32\linewidth]{#1} \\
\end{tabular}
}

\newcommand{\rowname}[1]{
\rotatebox[origin=c]{90}{\small {#1}}
}

\newcommand{\img}[1]{
\includegraphics[width=0.116\linewidth]{#1}
}

\definecolor{iccvblue}{rgb}{0.21,0.49,0.74}
\usepackage[pagebackref,breaklinks,colorlinks,allcolors=iccvblue]{hyperref}

\def\paperID{9630} 
\def\confName{ICCV}
\def\confYear{2025}

\usepackage{lipsum}

\title{PanoSplatt3R: Leveraging Perspective Pretraining for Generalized \\ Unposed Wide-Baseline Panorama Reconstruction}

\author{
Jiahui Ren\thanks{Equal contribution.} \quad 
Mochu Xiang\footnotemark[1] \quad 
Jiajun Zhu \quad 
Yuchao Dai\thanks{\ Corresponding author ({\tt daiyuchao@nwpu.edu.cn}). \\ This work was supported in part by the National Natural Science Foundation of China under Grants 62271410 and 12150007.} \\
School of Electronics and Information, Northwestern Polytechnical University and \\
Shaanxi Key Laboratory of Information Acquisition and Processing, Xi'an, Shaanxi, China
}

\begin{document}
\maketitle

\begin{abstract}
Wide-baseline panorama reconstruction has emerged as a highly effective and pivotal approach for not only achieving geometric reconstruction of the surrounding 3D environment, but also generating highly realistic and immersive novel views.
Although existing methods have shown remarkable performance across various benchmarks, they are predominantly reliant on accurate pose information. In real-world scenarios, the acquisition of precise pose often requires additional computational resources and is highly susceptible to noise. 
These limitations hinder the broad applicability and practicality of such methods.
In this paper, we present PanoSplatt3R, an unposed wide-baseline panorama reconstruction method. We extend and adapt the foundational reconstruction pretrainings from the perspective domain to the panoramic domain, thus enabling powerful generalization capabilities.
To ensure a seamless and efficient domain-transfer process, we introduce RoPE rolling that spans rolled coordinates in rotary positional embeddings across different attention heads, maintaining a minimal modification to RoPE's mechanism, while modeling the horizontal periodicity of panorama images.
Comprehensive experiments demonstrate that PanoSplatt3R, even in the absence of pose information, significantly outperforms current state-of-the-art methods. This superiority is evident in both the generation of high-quality novel views and the accuracy of depth estimation, thereby showcasing its great potential for practical applications.
Project page: \url{https://npucvr.github.io/PanoSplatt3R}
\end{abstract}    
\section{Introduction}
\label{sec:intro}

Panorama images capture 360$^\circ$ scene information from a single viewpoint, providing immersive visuals that can recreate the surrounding environment in full detail. This capability has found wide applications across various domains. In Virtual Reality (VR)~\cite{eiris2020desktop, brivio2021virtual}, Augmented Reality (AR)~\cite{liu2003augmented} and real estate~\cite{ramakrishnan2021habitat}, panorama images enhance user experiences by offering a more natural and engaging interaction with digital environments, enabling users to explore spaces remotely or in immersive settings. In autonomous driving and robotics~\cite{lemaire2007slam}, panorama images contribute to situational awareness, allowing systems to navigate complex environments with a broader context. These popular applications highlight the need for methods capable of accurately reconstructing entire 3D structures from panorama images while delivering photorealistic synthesis that allows users to freely explore and interact with the scene.
Given a pair of panorama images, generalized panorama reconstruction methods~\cite{zhang2024pansplat, chen2024splatter, chen2023panogrf} offer efficient and scalable solutions to this challenge. 

Existing methods address the challenge of reconstructing scene geometry and synthesizing novel views from two-view wide-baseline panorama images by constructing specialized cost volumes for panoramic data to estimate 360$^\circ$ depth maps. However, these approaches have two significant limitations. First, they require camera pose information as input, which not only demands additional computational resources but also introduces vulnerability to noise in real-world scenarios. Second, depth estimation from panoramic cost volumes constitutes a domain-specific problem, creating substantial barriers to the effective transfer of established knowledge from the perspective domain. For practical applications, there is a clear need for an unposed general method that demonstrates robust generalization capability across diverse scenarios.

Utilizing more than 8 million training samples, foundational stereo models in the perspective domain significantly advanced stereo matching, optical flow~\cite{weinzaepfel2023croco}, depth estimation, pose regression~\cite{dust3r}, visual localization and 3D reconstruction~\cite{mast3r}.
Given the broader knowledge base established in the perspective imagery, transfer learning from perspective to panoramic domains offers a promising approach to enhance model generalization abilities. Existing methods typically employ teacher-student frameworks that require frequent projection transformations between domains, which not only lacks efficiency but also continues to depend heavily on limited panoramic data. A fundamental challenge remains in how to effectively transfer general knowledge to this specialized domain. Development of a seamless and efficient knowledge transfer procedure would significantly advance this field, enabling more robust performance across diverse panoramic scenarios without the computational overhead of current approaches.

In this paper, we propose \textbf{PanoSplatt3R}, a novel approach for unposed wide-baseline panorama reconstruction. Our network adopts a vision transformer backbone and is trained to represent geometric structures from two viewpoints in a unified coordinate system, eliminating the need for camera poses~\cite{wang2025looprefine}. By leveraging a pretrained foundational model from the perspective domain, we transfer its strong generalization ability to work with panoramic data. Additionally, our framework incorporates specialized modeling to capture the geometric periodicity of panorama images. Enhanced with Gaussian Splatting for rendering, PanoSplatt3R faithfully reconstructs the 3D scene and generates photorealistic novel views.

We summarize our contributions as follows:
\begin{itemize}
    \item We introduce PanoSplatt3R, a novel unposed method for two-view wide-baseline panorama scene reconstruction and novel view synthesis, providing Gaussian Splatting based photorealistic renderings at novel viewpoints.
    \item We adapt a foundational stereo model for panorama images and demonstrate that general knowledge from the perspective domain can be effectively transferred to the panoramic domain, significantly improving wide-baseline panorama reconstruction.
    \item We propose RoPE rolling, a specialized variant of rotary positional embedding for panorama images, capturing geometric periodicity with minimal modifications to ensure seamless domain transfer.
    \item Extensive experiments show that PanoSplatt3R significantly outperforms state-of-the-art pose-dependent methods in both novel view synthesis quality and depth accuracy—even without known camera poses.
\end{itemize}

\begin{figure*}
\centering
\includegraphics[width=\linewidth]{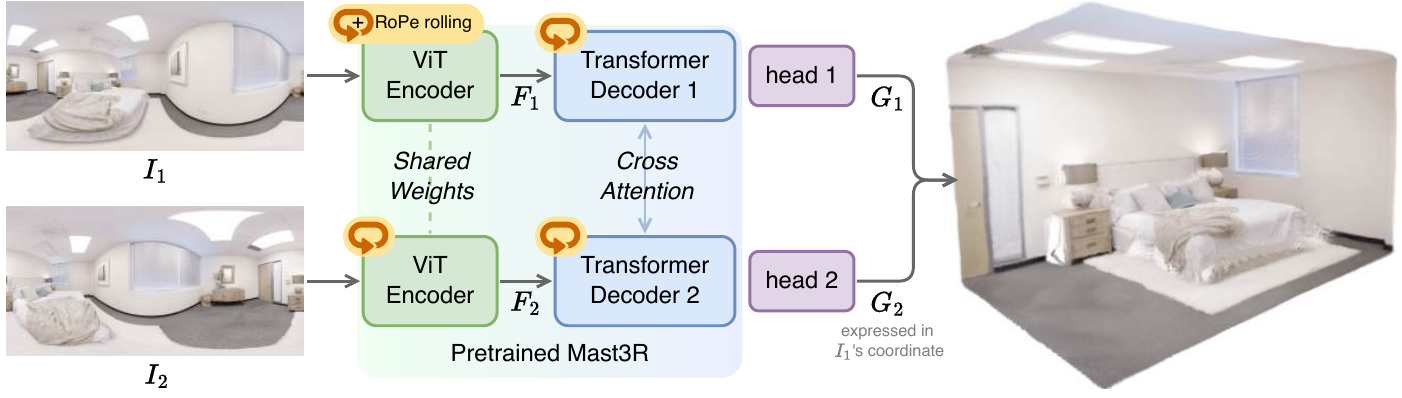}
\caption{\textbf{Overview of our proposed PanoSplatt3R.} Given two panorama images as input and without knowing their relative pose, our model reconstructs the entire 3D scene using Gaussian Splats. The estimated geometry can provide photorealistic renderings at novel view points. We removed two walls from the final reconstruction to better display the room.}
\label{fig:banner}
\end{figure*}

\section{Related Work}
\label{sec:related_work}

\subsection{Novel View Synthesis}

Novel View Synthesis (NVS) aims to generate visual representations of a scene from new viewpoints, given a set of known observations as input. This task typically involves either implicit scene reconstruction, such as Neural Radiance Fields (NeRF) \cite{mildenhall2020nerf, guo2023forward}, or explicit reconstruction methods, such as 3D Gaussian Splatting (3DGS) \cite{3dgs, lu20243d}.

Optimization-based NVS methods achieve high fidelity reconstruction through gradient-based scene fitting but are often computationally intensive. 
In contrast, generalized NVS methods are designed to operate with fewer input views and follow a feed-forward approach, enabling faster inference and better scalability to diverse scenes.

\textbf{Generalized Neural View Synthesis} methods learn scene priors from diverse multi-view data, enabling robust 3D scene reconstruction. These methods can either implicitly infer 3D geometry (e.g., pixelNeRF \cite{pixelnerf}) or explicitly express it using techniques like 3D Gaussian splats (e.g., pixelSplat \cite{pixelsplat}, MVSplat \cite{mvsplat}).

While explicit NVS methods benefit from better geometry estimation, this often becomes challenging in sparse-view scenarios, such as those involving only two images. In such cases, Dust3R \cite{dust3r} addresses these challenges by predicting two point maps in a shared coordinate frame, which not only enhances geometry estimation but also encodes camera intrinsics and relative pose information. Leveraging a diverse training dataset, Dust3R demonstrates strong generalization capabilities, enabling high-quality 3D scene reconstruction and improving 3D consistency in video generation tasks \cite{reconx, viewcrafter}.

Dust3R's advanced spatial understanding has been especially beneficial for generalized 3D Gaussian reconstruction, applying both at the object level \cite{vi3drm, freesplatter} and the scene level \cite{splatt3r, epipolar_free_3dgs, nopesplat, instantsplat, selfsplat, zerogs}. Its impact also extends to scene understanding \cite{large_spatial_model} and SLAM \cite{slam3r}, highlighting its versatility across various 3D vision applications.

To address challenges in processing longer sequences, several adaptations of the Dust3R architecture have been proposed \cite{fast3r, pref3r, spann3r, mv_dust3r_plus}. Furthermore, the pre-trained Dust3R model has shown effectiveness in dynamic scene reconstruction, as demonstrated in works like \cite{stereo4d, monst3r, align3r, cut3r}.

\subsection{3D Reconstruction from Panoramic Images}

Panorama images present unique challenges compared to standard perspective images due to their equirectangular projection. This projection creates a horizontal loop and encodes varying ray densities across rows, complicating feature extraction and holistic scene understanding. These distinctive properties necessitate specialized approaches to effectively process and analyze panoramic data.

To address these challenges, several approaches adopting convolutional networks introduce distortion-aware convolutions to expand the horizontal receptive field and mitigate projection-induced distortions \cite{tateno2018distortion, coors2018spherenet, de2018eliminating, zioulis2018omnidepth, eder2019pano, zhuang2022acdnet}. For transformer-based networks, solutions include deformable patch embeddings~\cite{zhang2022bending}, spherical token localization models \cite{shen2022panoformer}, and attention mechanisms incorporating spherical distance biases \cite{yun2023egformer}.

An alternative strategy involves projecting panorama images into multiple perspective images \cite{eder2020tangent}. Common techniques include tangent projection \cite{rey2022360monodepth, peng2023high, ai2023hrdfuse} and cubemap projection \cite{wang2020bifuse, jiang2021unifuse}. Since most models are pre-trained on perspective images, it enables better generalization while producing feature representations that are more compatible with standard networks. After feature extractions or direct applications of pre-trained models, the features of projected perspective images are fused back to the panoramic ones.

Other methods capture the intrinsic properties of panorama images through tailored representations. For instance, some approaches adopt column-wise \cite{sun2021hohonet} or panel-like structures \cite{yu2023panelnet}, some introduce spherical 3D point features to better model the geometry of panoramic scenes \cite{ai2024elite360d}.

Further advancements have been made for sparse panorama inputs, particularly with \textbf{two wide-baseline panorama} inputs. These methods typically construct a cost volume by using the extracted features and corresponding poses. PanoGRF \cite{chen2023panogrf} aggregates geometry from predicted depths and appearance features from the given panoramas to build the cost volume, but its computational cost is prohibitively high. Other methods enhance image quality by incorporating cube map images and depth features \cite{chen2024splatter}, or focus on high-resolution inputs and efficient training and rendering \cite{zhang2024pansplat}.

Panorama reconstruction methods can be broadly categorized into two main streams: varied wide baseline and fixed baseline. In the first setting, input panoramas are captured from arbitrary poses, introducing significant geometric variations and making the reconstruction process more challenging. In contrast, fixed-baseline methods~\cite{zhang2024pansplat} assume that the principal points of the input panoramas are aligned, and the goal is to synthesize an intermediate view. These methods inherently benefit from a predefined spatial relationship between views. A direct comparison would be unfair, as fixed-baseline methods operate within a controlled setup, but methods that handles varied wide baseline are more flexible in real-world applications. 
Our approach not only can handle varied-baseline inputs, but also eliminates the need for precise pose input, extending the application scenario of stereo panorama reconstruction.
\section{Method}
\label{sec:method}

\subsection{Overview}
Given two wide-baseline panoramic images $I_1$ and $I_2$, our goal is to densely reconstruct the scene using Gaussian Splatting. The model makes pixel-wise Gaussian parameter predictions $G_1$ and $G_2$, where each point is parameterized by $\{ \vb*{x}_i, \alpha_i, \vb*{r}_i, \vb*{s}_i, \vb*{c}_i\}_{i=1,...,2\times H\times W}$; specifically, the center location $\vb*{x}\in \mathbb R^3$, opacity $\alpha \in \mathbb R$, rotation $\vb*{r}\in \mathrm{SO}(3)$, scaling $\vb*{s}\in \mathbb R^3$ (For 2DGS~\cite{huang20242d}, scaling $\vb*{s}\in \mathbb R^2$) and color $\vb*{c}$ expressed with spherical harmonics.

An overview of our model is illustrated in~\autoref{fig:banner}.
Our model adopts a vision transformer backbone. Two panorama images $I_1$ and $I_2$ are first tokenized and feed into a shared encoder. Then the image features $F_1$ and $F_2$ from two branches attend each other in the decoder, which is composed of cross-attention blocks. Finally, the Gaussian parameters are predicted for each viewpoint. 

Reconstructing a scene from stereo panoramic images presents unique challenges. 
The remarkable generalization ability of foundational stereo models~\cite{dust3r, mast3r} does not directly apply to panoramic images, because they are predominantly trained on perspective data, leaving them ill-equipped to handle the inherent distortions and structural differences of panoramic inputs. Adaptation strategies are needed to bridge this domain gap and enable foundational stereo model to operate on panoramic inputs.

In the following sections, we first detail our approach, explaining how we achieve an unposed reconstruction in Section \ref{pose_free}.
Then we outline the key modifications that allow the model to overcome the challenges posed by panoramic imagery in Sec.~\ref{dust3r_pretain}. 
Next, we introduce a Gaussian parameter prediction head to enhance the visual realism of the reconstructions, as outlined in Sec.~\ref{gaussian_training}. 
A progressive training strategy is introduced in Sec.~\ref{model_training}, to ensure optimal performance in the panoramic context.

\subsection{Unposed Stereo Panorama Reconstruction}
\label{pose_free}
While previous methods~\cite{chen2023panogrf, chen2024splatter, zhang2024pansplat} tackle stereo panorama reconstruction by building cost volumes and finding correspondences, they rely on accurately known camera poses. However, in real-world scenarios, wide-baseline panoramic images are often captured without precise relative pose information, making these pipelines impractical. Moreover, estimating camera pose from two wide-baseline panoramic images is a challenging task. As a result, pose-dependent methods struggle when given inaccurate pose estimates, leading to degraded reconstruction quality.

Inspired by recent foundational stereo models~\cite{dust3r, mast3r}, we design our method to operate without requiring known camera poses. Our approach consists of three key components: (1) a shared ViT encoder that extracts features from both input images $I_1$ and $I_2$, (2) a transformer decoder that performs cross-attention between the encoded features $F_1$ and $F_2$, and (3) a Gaussian parameter prediction head that estimates the scene geometry. 
The model learns to establish correspondences between views through cross-attention and represents geometry in a unified coordinate system. Unlike cost volume-based methods, this approach removes the reliance on pose input, enabling wide-baseline panorama reconstruction without requiring known camera poses.

We build our backbone network upon the architecture of Dust3R~\cite{dust3r}. However, the original model fails to account for the horizontal periodicity of panoramic images, resulting in inconsistencies at the left and right boundaries. To overcome this limitation, we incorporate targeted enhancements, which we describe in detail in the next section.

\subsection{Adapting Perspective Models for Panorama}
\label{dust3r_pretain}

Positional embeddings enable vision transformers to sense the location of tokens, which is vital for precisely understanding 3D scenes.
We start from a brief review of positional embeddings. Currently, absolute positional embeddings are widely adopted in popular models \cite{vit, swin_transformer}, recent efforts to introduce relative positional embeddings have shown superior performance and greater flexibility \cite{heo2024rotary, lu2024fit}.
Rotary Positional Embedding (RoPE) \cite{su2024roformer} has been shown to be advantageous over absolute positional embeddings in tasks that rely on two-view correspondences \cite{weinzaepfel2023croco}, and is adopted in our backbone model~\cite{mast3r}. 

\begin{figure}[t]
    \centering
    \includegraphics[width=\linewidth]{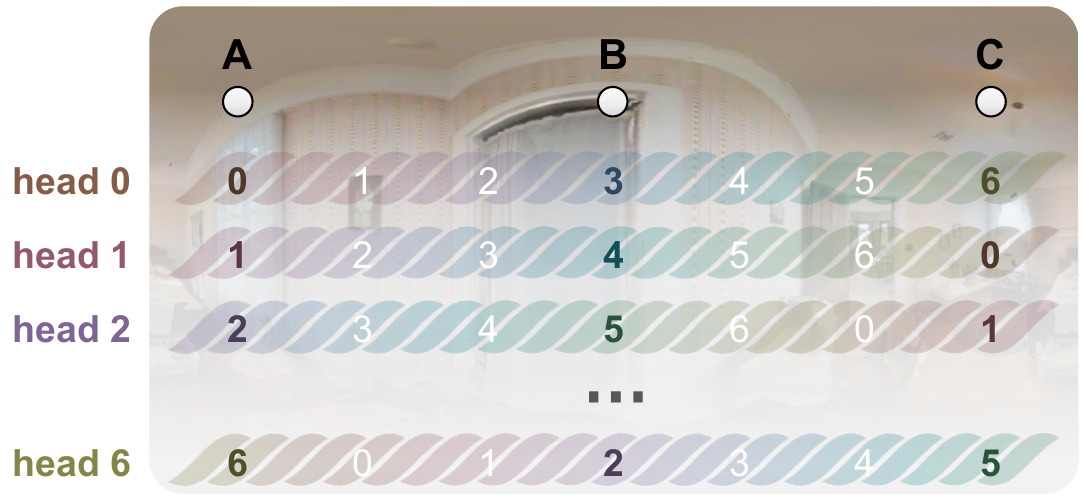}
    \caption{\textbf{Visual illustration of RoPE rolling.} 
    Multiple attention heads mark the location of A, B and C using different `RoPE's to avoid always distancing A and C in positional embeddings.}
    \label{fig:rope_rolling}
\end{figure}

As a relative positional embedding technique, RoPE applies a series of rotations at different frequencies to deep features, ensuring that the attention result after query and key transformations depends only on their relative spatial positions. When extending RoPE to 2D images, the feature vector is split into two halves: one undergoes rotation based on the $x$-axis position, while the other is rotated according to the $y$-axis position. For example, for a token located at $(p_n^x, p_n^y)$, the rotation matrices for feature dimension $t$ are:
\begin{align}
    \mathbf{R}(n, 2t) = e^{i\theta_tp_n^x}, \quad \mathbf{R}(n, 2t+1) = e^{i\theta_tp_n^y}.
\end{align}
In vision transformers, the rotation frequencies are typically set as $\theta_t = 100^{-t/(d_{head}/4)}$, for an attention head with $d_{head}$ channels. Using pixel locations allows RoPE to effectively capture spatial relationships in perspective images.

However, when applied to panorama images, this approach encounters fundamental challenges. Since panoramic images form a continuous horizontal loop, the left and right edges of the image are adjacent in the physical scene but are assigned the maximum possible horizontal separation in RoPE’s embedding space.
A straightforward approach might be to set the largest rotation angle in RoPE to $2\pi$ to enforce periodicity. However, this does not resolve the issue, as the frequency $\theta_t$ decreases with increasing feature dimension $t$. Intuitively, this means higher-dimensional features undergo slower rotations. Consequently, even if we increase $\theta_t$ by $2\pi$, only a small subset of feature dimensions would complete a full circular rotation, making it ineffective in modeling the continuous nature of panoramic images.

To address this, we propose a simple yet effective solution called \textit{\textbf{RoPE rolling}}. While a single RoPE application inherently follows a one-directional measurement without looping, it can be extended across multiple attention heads with varying offsets. For the $m^\text{th}$ attention head, we apply a horizontal roll to the $x$-coordinate by an offset of $Wm/M$, where $W$ is the feature map width and $M$ is the total number of attention heads. The resulting pixel location for the rotation matrices is formulated in \autoref{eq:rope_rolling}.
\begin{align}
\label{eq:rope_rolling}
    p_n^x \leftarrow (p_n^x + Wm/M) \mod W.
\end{align}
We apply RoPE rolling in both the encoder and decoder. \autoref{fig:rope_rolling} provides an intuitive illustration of this mechanism. This strategy minimally alters the original RoPE mechanism, preserving its compatibility with transfer learning while effectively modeling the periodic structure of panoramic images.

\subsection{Gaussian Parameter Prediction}
\label{gaussian_training}

We use the DPT head \cite{ranftl2021vision} to predict the Gaussian parameters $G_{1,2}$ for each view. Each view uses two DPT heads. The first one takes features from the transformer decoder and predicts the Gaussian center location $\vb*{x}$. The second head predicts remaining Gaussian parameters, utilizing features from the ViT decoder and the RGB image. The RGB image serves as a shortcut to directly incorporate texture information, which is essential for capturing fine details in 3D reconstruction \cite{nopesplat}. 

Since our model learns to predict the geometric structures in a common coordinate system, we fuse $G_1$ and $G_2$ by direct concatenation, then Gaussian rasterization kernels can render them at an arbitrary novel view.

\subsection{Model Training}
\label{model_training}

We employ a progressive training strategy where the model first learns to predict Gaussian centers before optimizing other Gaussian parameters with photometric loss. This staged optimization stabilizes training and improves convergence. During the first stage, we exclude the Gaussian parameter head responsible for predicting the remaining Gaussian parameters, focusing solely on learning accurate Gaussian centers. These centers ${\vb*{\hat x}}_{i}$ are supervised using ground truth 3D points ${\vb*{x}}_{i}$ obtained from depth projection. The loss for a valid pixel is formulated as the combination of $L_1$ and $L_2$ loss:
\begin{align}
    \mathcal{L}_{\text{stage1}} = \sum_{i=1}^{2\times H\times W} \left( \frac{1}{2} \| \vb*{x}_{i} - \hat{\vb*{x}}_{i} \| + \frac{1}{2} \| \vb*{x}_{i} - \hat{\vb*{x}}_{i}\|^2 \right ).
\end{align}
To supervise the model to express points of $I_2$ in the coordinate of $I_1$, we project the points in $I_2$ to 3D using its ground truth depth $D_2$, then transform these points to $I_1$'s coordinate using the relative pose between them.

In the second stage, we supervise all Gaussian parameters at a novel view. To provide holistic supervision of the entire reconstructed scene, we use the cubemap projection to capture the surrounding scene of a viewpoint. Specifically, we render 6 images with $90^\circ$ field of view (FOV) at the same position but faces 6 different directions. Utilizing the rendered image $\hat I$ and depth $\hat D$, the full model is trained under both image and depth losses. For image supervision, we use the mean squared error (MSE) and LPIPS loss, with weights of 1.0 and 0.05, respectively. For depth supervision, we use $L_2$ loss with a weight of 0.08. 
\begin{align}
    \mathcal{L_\text{stage2}} &= \sum_{j=1}^{6} \left[
    \left\| I_j - \hat I_j \right\|_2 
    + 0.08\left\| D_j - \hat D_j \right\|_2 \right .
    \nonumber \\  &
    \left. + 0.05 \cdot \mathrm{LPIPS} (I_j, \hat I_j) \right],
\end{align}
where $I_j$ and $D_j$ are the ground truth image and depth for the $j^\text{th}$ cube map view at a novel viewpoint.
\section{Experiments}
\label{sec:experiments}

\begin{table*}[!t]
\centering
\caption{\textbf{Quantitative comparisons on wide-baseline panorama reconstructions.} We compare our model with methods that require accurate known pose, including those with cube map inputs and panorama inputs. Our model, even in the absence of camera pose, presents the best performance in both novel view synthesis quality and depth estimation accuracy. Models with \textsuperscript{\dag} are trained on HM3D by \cite{chen2024splatter}.}
\setlength{\tabcolsep}{4pt}
\resizebox{\textwidth}{!}{
\begin{tabular}{r | l | ccc|ccc | ccc|ccc} \toprule
\multirow{2}{*}{\begin{tabular}{@{}r@{}}\textbf{Problem} \\ \textbf{Setting}\end{tabular}}  & & \multicolumn{6}{c|}{\textbf{HM3D}} & \multicolumn{6}{c}{\textbf{Replica}} \\
 & \textbf{Method} &
{\small PSNR$\uparrow$} & {\small SSIM$\uparrow$} & {\small LPIPS$\downarrow$} &
{\small Rel$\downarrow$} & {\small RMS$\downarrow$} & {\small $\delta_{1}\uparrow$} &

{\small PSNR$\uparrow$} & {\small SSIM$\uparrow$} & {\small LPIPS$\downarrow$} &
{\small Rel$\downarrow$} & {\small RMS$\downarrow$} & {\small $\delta_{1}\uparrow$} 
\\ \midrule

\multirow{3}{*}{\begin{tabular}{@{}r@{}}\textit{posed} \\ \textit{perspective}\end{tabular}} &
HiSplat \cite{tang2024hisplat} &
    17.358 & 0.620 & 0.486  &
    0.678  & 0.907 & 29.990 &
    17.157 & 0.642 & 0.417  &
    0.740  & 1.025 & 25.824 \\
 & MVSplat \cite{mvsplat} &
    18.015 & 0.627 & 0.511  &
    1.186  & 1.367 & 24.884 &
    18.005 & 0.631 & 0.512  &
    1.423  & 1.798 & 21.229 \\

 & DepthSplat \cite{xu2024depthsplat} &
    18.487 & 0.652 & 0.456  &
    0.570  & 0.693 & 37.791 &
    19.369 & 0.732 & 0.334  &
    0.373  & 0.550 & 50.364 \\
\midrule

\multirow{3}{*}{\begin{tabular}{@{}r@{}}\textit{posed} \\ \textit{360$^\circ$}\end{tabular}} &
PanoGRF\cite{chen2023panogrf} &
    25.863 & 0.810 & 0.304  &
    0.195  & 0.308 & 85.168 &
    27.920 & 0.892 & 0.171  &
    0.128  & 0.258 & 88.057 \\
    
 & MVSplat\textsuperscript{\dag} \cite{mvsplat} &
    27.227 & 0.852 &  0.175 &
     0.094 & 0.270 & 90.505 &
    28.399 & 0.908 &  0.115 &
     0.088 & 0.247 & 89.913 \\
    
 & Splatter-360 \cite{chen2024splatter} &
    28.308 & \textbf{0.875} & 0.154  &
     0.078 & 0.224 & 93.627 &
    29.888 & 0.924 & 0.097  &
     0.063 & 0.197 & 94.572 \\ \midrule
    
{\textit{unposed 360$^\circ$}} &PanoSplatt3R &
    \textbf{28.938} & 0.869 &  \textbf{0.147} &
     \textbf{0.044} & \textbf{0.147} & \textbf{97.654} &
    \textbf{31.522} & \textbf{0.935} &  \textbf{0.082} &
     \textbf{0.034} & \textbf{0.126} & \textbf{98.115} \\
\bottomrule
\end{tabular}
} 
\label{tab:comparison}
\end{table*}

\subsection{Experimental Settings}
\label{sec:exp_setting}

\begin{table*}
\centering
\begin{minipage}[t]{0.20\textwidth}
\centering
\caption{\textbf{Relative pose estimation comparison.} Results are measured on the Replica dataset and metrics are averaged.}
\vspace{2pt}
\resizebox{\textwidth}{!}{
\setlength{\tabcolsep}{3pt}
\begin{tabular}{l|c c} \toprule
    \textbf{Method} & RRA & RTA \\ \midrule
    SIFT+8PA & 20.53$^\circ$ & 14.07$^\circ$\\
    Ours+PnP & 0.541$^\circ$ & 0.584$^\circ$\\
    Ours+8PA & \textbf{0.477}$^\circ$ & \textbf{0.239}$^\circ$\\
    \bottomrule
\end{tabular}
\label{tab:pose_est}
} 
\end{minipage}
\hfill
\begin{minipage}[t]{0.78\textwidth}
\centering
\caption{\textbf{Quantitative comparisons on unposed pipelines.} We use SIFT to get point matches and use the 8-points algorithm (8PA) to estimate the relative pose, then apply posed methods with it. Current methods are vulnerable towards pose inaccuracy. Our model does not require pose input, performs significantly better.}
\resizebox{\textwidth}{!}{
\setlength{\tabcolsep}{3pt}
\begin{tabular}{l | ccc|ccc | ccc|ccc}
\toprule
 & \multicolumn{6}{c|}{\textbf{HM3D}} & \multicolumn{6}{c}{\textbf{Replica}} \\
\textbf{Method} &
{\small PSNR$\uparrow$}  & {\small SSIM$\uparrow$}  & {\small LPIPS$\downarrow$} &
{\small Rel$\downarrow$} & {\small RMS$\downarrow$} & {\small $\delta_{1}\uparrow$} &
{\small PSNR$\uparrow$}  & {\small SSIM$\uparrow$}  & {\small LPIPS$\downarrow$} &
{\small Rel$\downarrow$} & {\small RMS$\downarrow$} & {\small $\delta_{1}\uparrow$} 
\\ \midrule

8PA+PanoGRF \cite{chen2023panogrf} &
    18.522 & 0.660 &  0.488 &
     0.396 & 0.658 & 54.968 &
    17.733 & 0.699 &  0.430 &
     0.317 & 0.653 & 54.422 \\
    
8PA+MVSplat\textsuperscript{\dag} \cite{mvsplat} &
    16.386 & 0.581 &  0.512 &
     0.429 & 0.831 & 40.747 &
    14.346 & 0.607 &  0.465 &
     0.454 & 0.942 & 31.534 \\
    
8PA+Splatter-360 \cite{chen2024splatter} &
    14.765 & 0.537 &  0.496 &
     0.448 & 0.858 & 40.225 &
    12.916 & 0.570 &  0.463 &
     0.436 & 0.920 & 34.223 \\ \midrule

PanoSplatt3R &
    \textbf{28.938} & \textbf{0.869} &  \textbf{0.147} &
     \textbf{0.044} & \textbf{0.147} & \textbf{97.654} &
    \textbf{31.522} & \textbf{0.935} &  \textbf{0.082} &
     \textbf{0.034} & \textbf{0.126} & \textbf{98.115} \\
\bottomrule
\end{tabular}
\label{tab:all_unposed}
} 
\end{minipage}
\end{table*}

\noindent\textbf{Datasets.} To compare with existing methods, we evaluate PanoSplatt3R on two synthetic datasets: HM3D \cite{ramakrishnan2021habitat} and Replica \cite{straub2019replica}. We follow the data split of Splatter-360 \cite{chen2024splatter} to render panoramic videos with AI-Habitat simulation tool \cite{savva2019habitat} by sampling random trajectories, and train on HM3D. All panoramic images are at a resolution of 512$\times$1024. Since the data preparation process involves random sampling, reported metrics are not identical as in previous works~\cite{chen2024splatter}, but are still highly consistent with them.

\noindent\textbf{Comparing Methods.} We adhere to the evaluation protocol established in Splatter-360 and present performance comparisons with wide-baseline panorama reconstruction methods including Splatter-360 \cite{chen2024splatter} and PanoGRF \cite{chen2023panogrf}, along with perspective methods that takes cubemap images, including HiSplat \cite{tang2024hisplat}, MVSplat \cite{mvsplat} and DepthSplat \cite{xu2024depthsplat}.

\noindent\textbf{Evaluation Metrics.} 
We evaluate the quality of novel view synthesis using PSNR, SSIM, and LPIPS. Additionally, we compare the rendered depth map errors to assess the accuracy of 3D reconstruction, including absolute relative error (AbsRel), root mean squared error (RMSE) and accuracy under threshold $\delta_1$($\delta < 1.25$).

\noindent\textbf{Scale Recovery.} 
While we supervise the model using data in metric units, recovering metric geometry from two viewpoints with an unknown baseline distance remains an ill-posed problem. This results in misalignment between the ground truth view and the estimation view rendered with a metric pose.

To address this, we adopt a straightforward strategy by applying a uniform scaling to the model’s estimated geometry. Specifically, we begin by estimating the relative pose of the second input view and solve a Perspective-n-Point (PnP)~\cite{lepetit2009ep} problem. Since the geometry of the second viewpoint is expressed in the coordinate frame of the first viewpoint, we directly construct the 3D-2D correspondences using the point map of the second view and its 2D image coordinates. Once the relative pose, denoted as $\mathbf{\hat P}_2 = \left[\mathbf{\hat R} \left| {\vb*{\hat t}} \right.\right]$, is obtained, we compute a scaling factor $\lambda$ based on the ratio of the norms of the translation vectors: $\lambda = |\vb*{t}|/|\vb*{\hat t}|$, where $\vb*{t}$ is the translation from the ground truth pose and $\vb*{\hat t}$ is the estimated translation.

We apply the scaling to both the location and scale of the 3D Gaussians:
\(
    \vb*{x} \leftarrow \lambda \vb*{x}, \quad \vb*{s} \leftarrow \lambda \vb*{s} .
\)
Please note that this process only scales the reconstruction up to a global factor, and does not affect the quality of the reconstruction; rather, it is applied to eliminate scale ambiguity, allowing the photometric indicators to more accurately reflect the true reconstruction quality.

\subsection{Implementation Details}

PanoSplatt3R is implemented in Pytorch, with a 2DGS \cite{huang20242d} renderer implemented in CUDA. We initialize our model using pre-trained weights of Mast3R \cite{mast3r}. 
We train the model with 50k steps in stage 1 and 100k steps in stage 2.
To improve efficiency and reduce GPU memory usage, we replace the original attention mechanism with xFormers' memory-efficient attention \cite{xFormers2022} and utilize auto mixed precision training. Experiments are conducted on 8 NVIDIA L40 48G GPUs.

\subsection{Experimental Results}

We present performance comparisons in~\autoref{tab:comparison}. As an unposed approach, PanoSplatt3R significantly outperforms existing pose-dependent methods in both novel view synthesis quality and depth accuracy. Visualization results in~\autoref{fig:depth} further highlight its effectiveness, showcasing photorealistic novel views and highly accurate 3D structures. Please find more visualization results in the supplementary.

\subsection{Further discussions}

\subsubsection{Applying posed methods to the unposed setting.}

Current wide-baseline panorama reconstruction methods rely on accurate pose information, which is often difficult to obtain in real-world applications. To evaluate their robustness, we first conduct an experiment on the relative pose estimation on wide-baseline panorama images, results are displayed in~\autoref{tab:pose_est}, using inputs from the Replica~\cite{straub2019replica} dataset. The relative pose derived using SIFT~\cite{sift} and 8-point algorithm~\cite{hartley1997defense} with RANSAC~\cite{fischler1981ransac} (SIFT+8PA) is highly inaccurate measured by Relative Rotation Angle (RRA) and Relative Translation Angle (RTA). 

In contrast, our model's PnP-based relative pose estimation (as detailed in Section~\ref{sec:exp_setting}) is significantly more accurate. 
We also use the estimated 3D points from two images to find 2D correspondences and apply 8-point algorithm to recover the relative pose, the results are even more accurate, the relative rotation and translation angle are both $<0.5^\circ$.

We further conduct experiments using estimated relative poses (with translations properly scaled) and present the results in~\autoref{tab:all_unposed}. Since estimating accurate pose from wide-baseline panoramic images is itself a challenging problem, pose-dependent methods are highly sensitive to pose inaccuracies, resulting in degraded performance. This underscores the need for a pose-free wide-baseline panorama reconstruction approach. 

\subsubsection{Comparison with fixed baseline methods.}

\begin{table*}[!ht]
\caption{
\textbf{Quantitative comparisons with fixed baseline methods.} Results are measured on the novel view panorama synthesis. All models above the horizontal line are trained on Matterport3D with a baseline of 1.0 meter, their results are measured by \cite{zhang2024pansplat}.}
\setlength\tabcolsep{2pt}
\centering
\resizebox*{1.0\linewidth}{!}{
\begin{tabular}{l|ccc|ccc|ccc|ccc|ccc}
\toprule
 &
\multicolumn{3}{c|}{\textbf{Matterport3D} (1.0m)} &
\multicolumn{3}{c|}{\textbf{Matterport3D} (1.5m)} &
\multicolumn{3}{c|}{\textbf{Matterport3D} (2.0m)} &
\multicolumn{3}{c|}{\textbf{Replica} (1.0m)} &
\multicolumn{3}{c}{\textbf{Residential} ($\sim$0.3m)} \\
\textbf{Method} &
{\small WS-PSNR$\uparrow$} & {\small SSIM$\uparrow$} & {\small LPIPS$\downarrow$} & 
{\small WS-PSNR$\uparrow$} & {\small SSIM$\uparrow$} & {\small LPIPS$\downarrow$} &
{\small WS-PSNR$\uparrow$} & {\small SSIM$\uparrow$} & {\small LPIPS$\downarrow$} &
{\small WS-PSNR$\uparrow$} & {\small SSIM$\uparrow$} & {\small LPIPS$\downarrow$} &
{\small WS-PSNR$\uparrow$} & {\small SSIM$\uparrow$} & {\small LPIPS$\downarrow$} \\ \midrule
S-NeRF~\cite{mildenhall2020nerf} &
15.25 & 0.579 & 0.546 &
14.16 & 0.563 & 0.580 &
13.13 & 0.523 & 0.607 &
16.10 & 0.723 & 0.443 &
22.47 & 0.741 & 0.435 \\

IBRNet~\cite{wang2021ibrnet} &
25.72 & 0.855 & 0.258 &
21.69 & 0.751 & 0.382 &
20.04 & 0.706 & 0.431 &
22.65 & 0.854 & 0.291 &
22.47 & 0.735 & 0.498 \\

NeuRay~\cite{liu2022neural} &
24.92 & 0.832 & 0.260 &
21.92 & 0.766 & 0.347 &
19.85 & 0.715 & 0.407 &
25.90 & 0.899 & 0.187 &
22.38 & 0.753 & 0.427 \\

PanoGRF~\cite{chen2023panogrf} & 
27.12 & 0.876 & 0.195 &
23.38 & 0.811 & 0.282 &
20.96 & 0.761 & 0.352 &
29.22 & 0.937 & 0.134 &
31.03 & 0.909 & 0.207 \\

MVSplat~\cite{mvsplat} & 
29.29 & 0.912 & 0.105 &
22.51 & 0.807 & 0.230 &
13.38 & 0.595 & 0.554 &
31.25 & 0.958 & 0.059 &
31.32 & 0.906 & 0.200 \\ 

PanSplat~\cite{zhang2024pansplat} &
\textbf{30.01} & \textbf{0.931} & \textbf{0.091} &
24.76 & \textbf{0.849} & \textbf{0.181} &
21.19 & 0.777 & 0.265 &
\textbf{31.67} & \textbf{0.962} & \textbf{0.069} &
\textbf{31.36} & \textbf{0.917} & \textbf{0.172} \\ \midrule

Splatter-360\cite{chen2024splatter} &
21.97 & 0.750 & 0.273 &
19.91 & 0.697 & 0.354 &
18.50 & 0.659 & 0.395 &
24.56 & 0.875 & 0.191 &
22.95 & 0.735 & 0.366 \\

PanoSplatt3R & 
27.47 & 0.876 & 0.145 &
\textbf{25.56} & 0.844 & 0.185 &
\textbf{23.65} & \textbf{0.804} & \textbf{0.241} &
28.64 & 0.931 & 0.106 &
28.86 & 0.852 & 0.264\\ \bottomrule
\end{tabular}
} 
\label{tab:fixed_baseline}
\end{table*}

\begin{figure*}[t]
\resizebox{\textwidth}{!}{
\centering
\setlength{\tabcolsep}{0.5pt}
\begin{tabular}{c@{\hspace{-2.5pt}}c c@{\hspace{-2.5pt}}c c@{\hspace{-2.5pt}}c | @{\hspace{2.5pt}} c@{\hspace{-2.5pt}}c}

\multicolumn{2}{c}{\small MVSplat\textsuperscript{\dag} \cite{mvsplat}} & 
\multicolumn{2}{c}{\small Splatter-360\cite{chen2024splatter}} &
\multicolumn{2}{c|@{\hspace{2.5pt}}}{\small PanoSplatt3R} & 
\multicolumn{2}{c}{\small GT} \\

\img{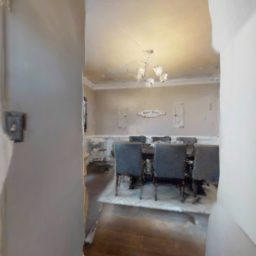} &
\img{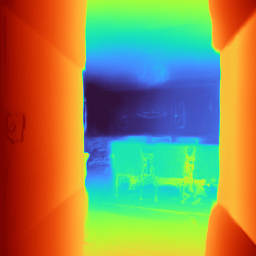} &
\img{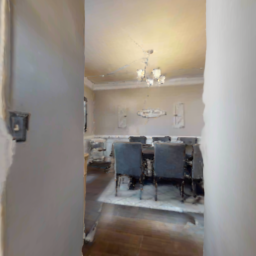} &
\img{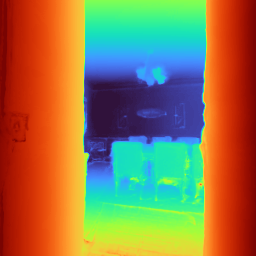} &
\img{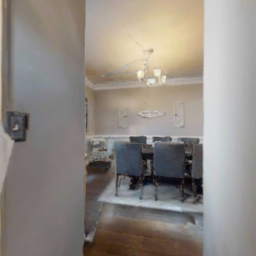} &
\img{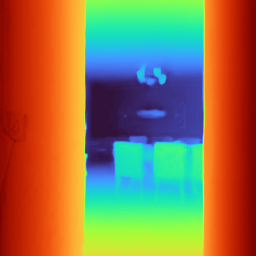} &
\img{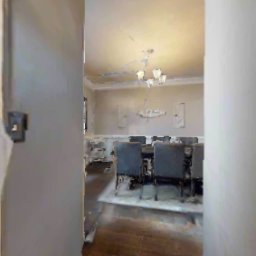} &
\img{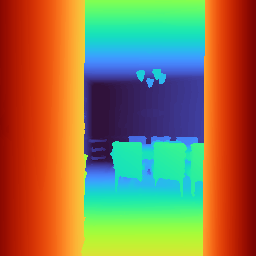} \\

\img{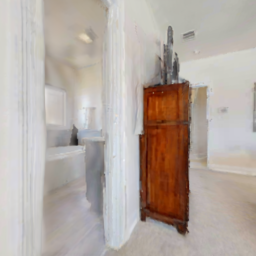} &
\img{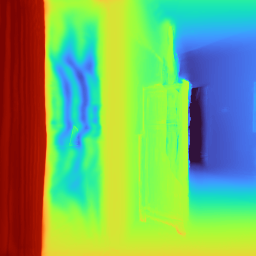} &
\img{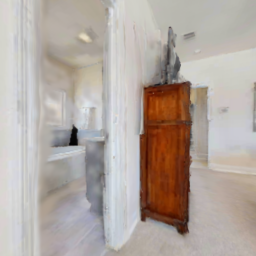} &
\img{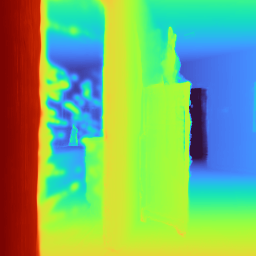} &
\img{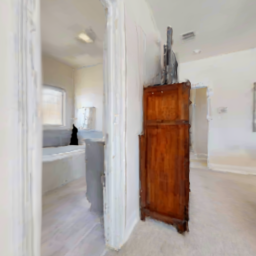} &
\img{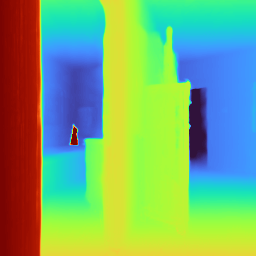} &
\img{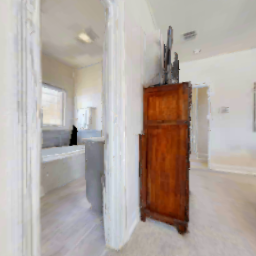} &
\img{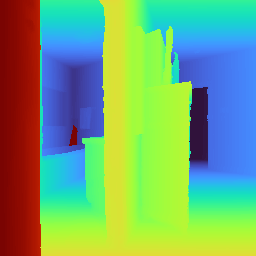} \\



\img{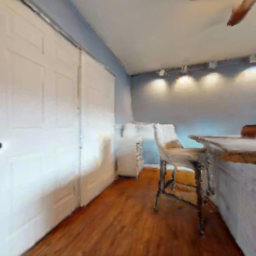} &
\img{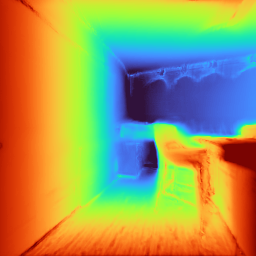} &
\img{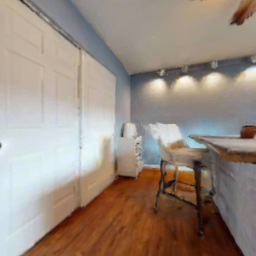} &
\img{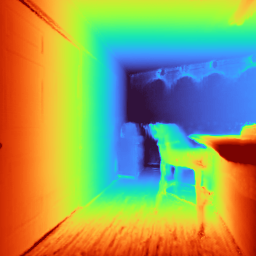} &
\img{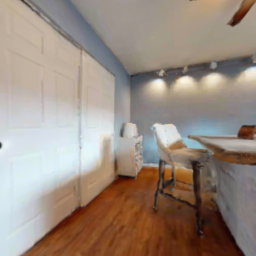} &
\img{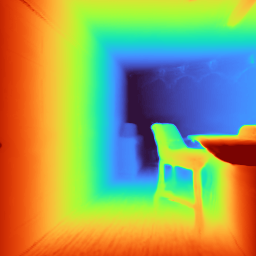} &
\img{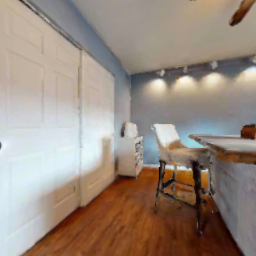} &
\img{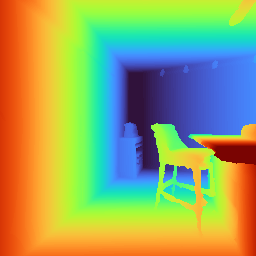} \\

\img{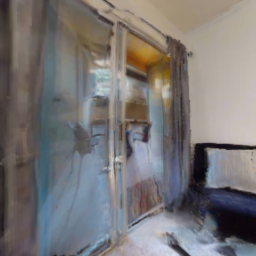} &
\img{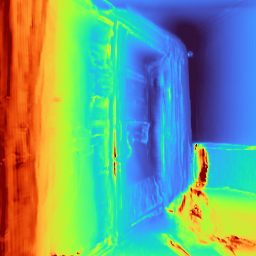} &
\img{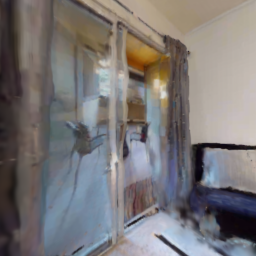} &
\img{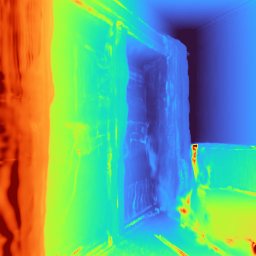} &
\img{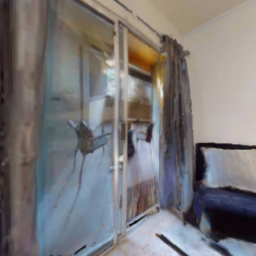} &
\img{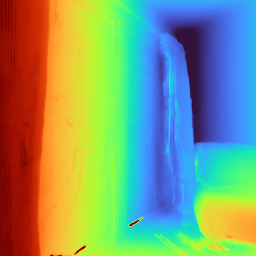} &
\img{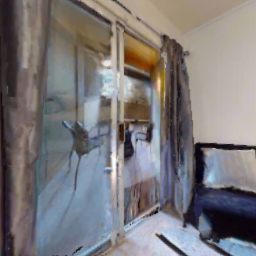} &
\img{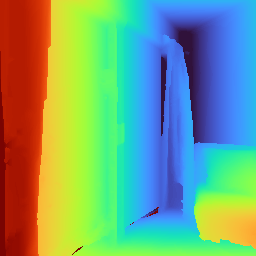} \\

\img{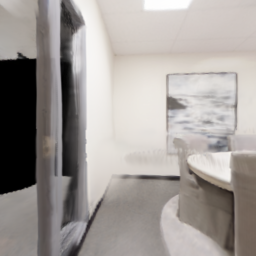} &
\img{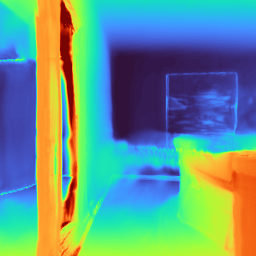} &
\img{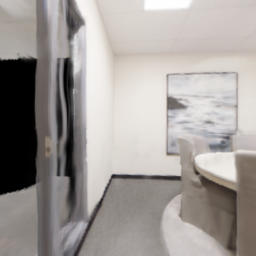} &
\img{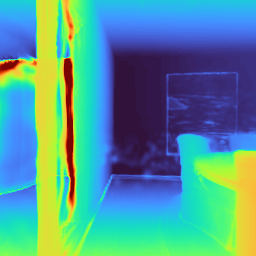} &
\img{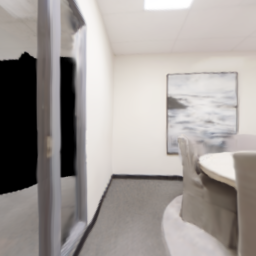} &
\img{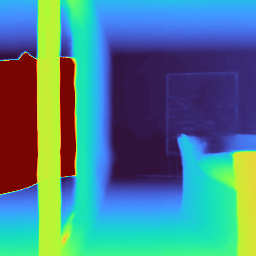} &
\img{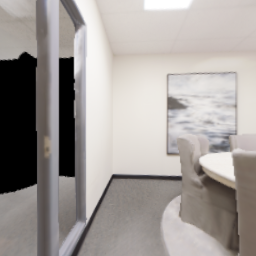} &
\img{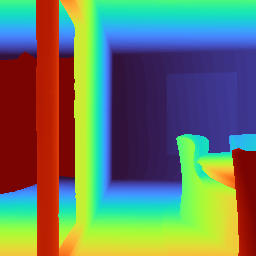} \\

\img{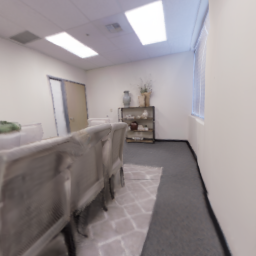} &
\img{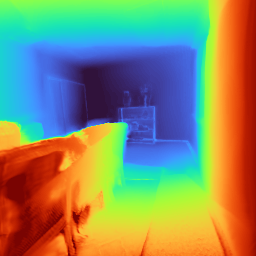} &
\img{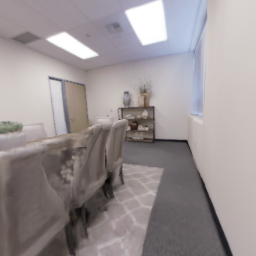} &
\img{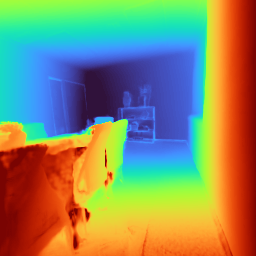} &
\img{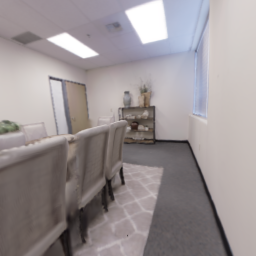} &
\img{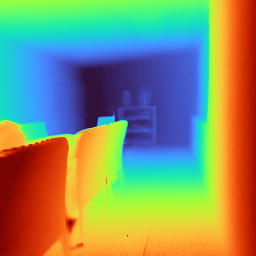} &
\img{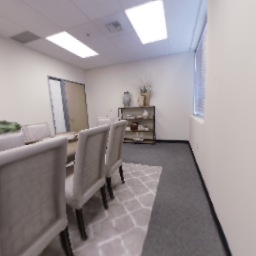} &
\img{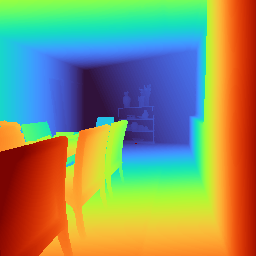} \\

\end{tabular}
} 
\caption{\textbf{Visual Comparison on reconstructed image} (above) \textbf{and depth} (below). Each row comes from different pairs of input. 
}
\label{fig:depth}
\end{figure*}

\begin{table*}
\centering
\caption{\textbf{Ablation studies on method details (model components and design principles).} }
\setlength{\tabcolsep}{4pt}
\resizebox{\textwidth}{!}{
\begin{tabular}{l | ccc|ccc|ccc|ccc}
\toprule
 & \multicolumn{6}{c|}{\textbf{HM3D}} & \multicolumn{6}{c}{\textbf{Replica}} \\
\textbf{Configuration} &
{\small PSNR$\uparrow$}  & {\small SSIM$\uparrow$}  & {\small LPIPS$\downarrow$} &
{\small Rel$\downarrow$} & {\small RMS$\downarrow$} & {\small $\delta_{1}\uparrow$} &
{\small PSNR$\uparrow$}  & {\small SSIM$\uparrow$}  & {\small LPIPS$\downarrow$} &
{\small Rel$\downarrow$} & {\small RMS$\downarrow$} & {\small $\delta_{1}\uparrow$} 

\\ \midrule

a. $\times$ RoPE rolling & 
27.916 & 0.843 &  0.165 &
 0.053 & 0.161 & 97.247 &
27.852 & 0.877 &  0.122 &
 0.067 & 0.178 & 96.398 \\

b. $\rightarrow$ RoPE 2$\pi$  & 
26.849 & 0.818 &  0.187 &
 0.067 & 0.181 & 96.150 &
26.695 & 0.862 &  0.148 &
 0.084 & 0.213 & 93.408 \\

c. $\times$ Mast3R weight &
23.331 & 0.731 &  0.281 &
 0.157 & 0.321 & 85.006 &
23.673 & 0.815 &  0.212 &
 0.172 & 0.364 & 79.310 \\

d. $\times$ Progressive training&
27.217 & 0.827 &  0.182 &
 0.060 & 0.173 & 96.707 &
27.918 & 0.882 &  0.128 &
 0.067 & 0.179 & 95.255 \\

e. $\times$ 2DGS $\rightarrow$ 3DGS &
28.049 & 0.846 &  0.162 &
 0.051 & 0.160 & 97.291 &
28.069 & 0.880 &  0.118 &
 0.067 & 0.174 & 96.645 \\ 
  
 \midrule
Full PanoSplatt3R &
28.200 & 0.848 &  0.160 &
 0.051 & 0.157 & 97.397 &
28.181 & 0.882 &  0.117 &
 0.063 & 0.168 & 97.055  \\

~~~~$+$ Scale &
28.938 & 0.869 & 0.147 &
0.044  & 0.147 & 97.654 &
31.522 & 0.935 & 0.082 &
0.034  & 0.126 & 98.115 \\
\bottomrule
\end{tabular}
} 
\label{tab:model_ablation}
\end{table*}

\begin{figure}[t]
\centering
\setlength{\tabcolsep}{1pt}
\resizebox{\linewidth}{!}{
\begin{tabular}{@{}c c c@{}}
\includegraphics[width=0.49\linewidth]{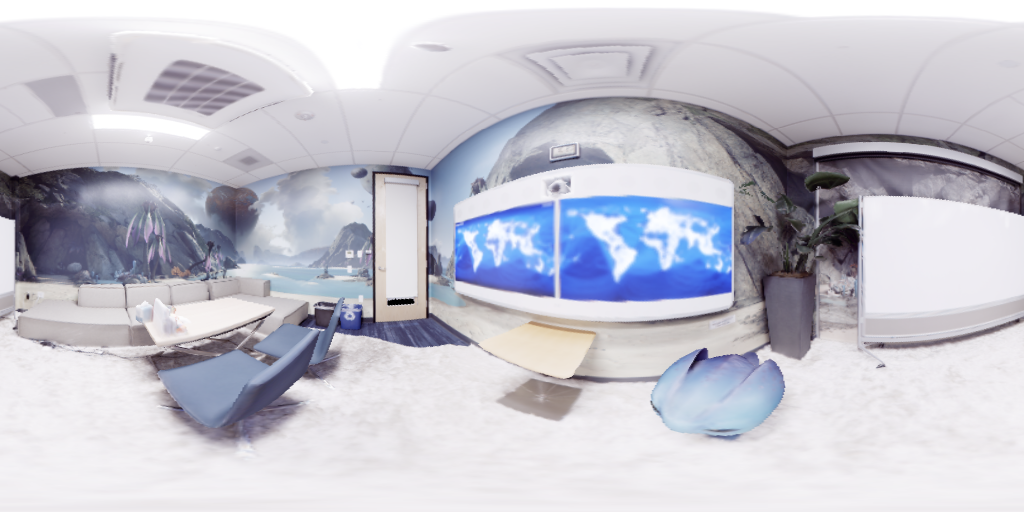} &
\includegraphics[width=0.24\linewidth]{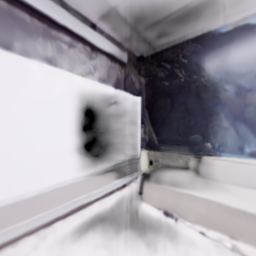} &
\includegraphics[width=0.24\linewidth]{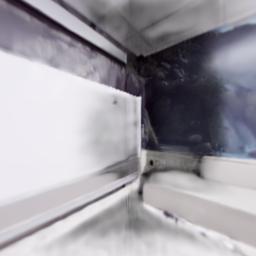} \\
{\small $I_1$ in input views} & {\small w/o RoPE Rolling} & {\small Full model} \\
\end{tabular}
} 
\caption{Model without RoPE Rolling estimates broken geometry at panorama image boundaries, while the full PanoSplatt3R reconstructs the wall without a hole.}
\label{fig:visual_wo_rope_rolling}
\end{figure}

We compare our model against methods trained and evaluated under a fixed-baseline setting, as shown in~\autoref{tab:fixed_baseline}. Here, the principal axes of the two panoramic images are aligned, which makes the stereo reconstruction task significantly easier for posed methods. Since these methods are trained under different assumptions and data splits, direct comparisons are not entirely fair. We add Splatter-360~\cite{chen2024splatter}, who is trained on the same data as ours for a fair comparison. Our model outperforms Splatter-360 by a large margin, showcasing its strong generalization ability.

Despite not being trained on Matterport3D~\cite{ramakrishnan2021habitat} and being designed to handle more challenging and unconstrained inputs, our model achieves comparable performance to these methods with narrow-baseline panorama images on Replica~\cite{straub2019replica} and Residential~\cite{habtegebrial2022somsi}, and even outperform them in the wider-baseline setting. Moreover, as the baseline distance increases, our model performance stably decreases, while fixed baseline methods face a sharp decline.

\subsection{Ablation Studies}
In this section, we conduct various ablation studies to analyze the effectiveness of our model components and design principles. The results are reported in~\autoref{tab:model_ablation}.

\noindent\textbf{Important of perspective pre-training.}
Configuration c) (Cfg.c) adopts a random initialization while our full model uses pretrained weights of Mast3R. There is a significant performance gap between Cfg.c and full model, demonstrating the effectiveness of Dust3R's and Mast3R's perspective pre-training, while highlighting the necessity of we adopting a transfer learning paradigm. 

\noindent\textbf{Effectiveness of RoPE rolling.} 
Configuration a) uses the standard RoPE implementation from Dust3R and Mast3R, which performs worse than the full model. This indicates that RoPE rolling is crucial in improving geometric predictions for panoramic data. To better illustrate its impact, we visualize the Gaussian prediction $G_1$ separately. As shown in~\autoref{fig:visual_wo_rope_rolling}, the input panorama image $I_1$ contains a wall spanning across its right and left edges. Without RoPE rolling, the model fails to capture the continuity of the wall’s geometry, leaving a gap in the final reconstruction. In contrast, our full model, equipped with RoPE rolling, successfully estimates the complete and continuous structure.

As explained in Section~\ref{dust3r_pretain}, changing the maximum rotating angle to $2\pi$ does not yield a horizontally periodical positional encoding. What's worse, it changes the RoPE mechanism, hindering the knowledge transfer process. Its effect is further validated in Cfg.b.

\noindent\textbf{Importance of progressive training of the model.}
Configuration d) skips stage 1 and directly trains the model in stage 2. For a fair comparison, we train it for 150K steps—the total number of steps across both stages in our full model. In the early training phase, the model produces random point predictions, and each Gaussian point has a limited receptive field for backpropagating gradients after being rasterized to 2D. Directly optimizing with photometric loss at this stage hinders learning. Instead, we first stabilize the model’s 3D point predictions before introducing photometric supervision, leading to a more effective training process.

\noindent\textbf{What if model directly regresses metric predictions?}
We apply scaling to the final reconstruction to resolve the scale ambiguity inherent in unposed inputs. For completeness, we also evaluate our model without this scaling step, denoted as ``Full PanoSplatt3R''. Even without scaling, the model achieves leading performance, demonstrating its strong generalization ability in unposed wide-baseline panorama reconstruction.

Furthermore, Cfg.e in~\autoref{tab:model_ablation} implies that using 2DGS can bring superior performance than using 3DGS.

\section{Conclusion}
\label{sec:conclusion}

In this paper, we proposed PanoSplatt3R, a novel framework for reconstruction and view synthesis that operates directly on unposed wide-baseline panoramic images.
Our approach facilitates seamless knowledge transfer from the perspective domain to the panoramic domain, improving the generalization capabilities for this challenging task. 
Unlike existing methods that require accurate poses, PanoSplatt3R achieves superior performance without relying on camera poses, demonstrating its strong capabilities.
By eliminating the requirement of camera poses, PanoSplatt3R enables broader application of wide-baseline panorama reconstruction in diverse practical applications.

\clearpage
{
    \small
    \bibliographystyle{ieeenat_fullname}
    \bibliography{main}
}

\clearpage
\appendix
\addcontentsline{toc}{section}{Supplementary Material}
\maketitlesupplementary

\section{Experimental Details}

\textbf{Training Details.} 

Our model is trained on random pairs of panoramic images sampled from rendered HM3D video, with frame intervals evenly sampled between 75 and 100. The frame used for supervision is randomly selected between the two frames. 

We employ the AdamW optimizer \cite{loshchilov2017decoupled} for training, with an initial learning rate of $2 \times 10^{-5}$ in the first stage. In the second stage, the initial learning rate for the backbone is set to $2 \times 10^{-5}$, while the learning rate for the remaining Gaussian parameter head is set to $2 \times 10^{-4}$. A warm-up strategy is applied for both stages over 2k steps. The final learning rate will decay to $1/10$ of the original value.

\noindent\textbf{Dataset Details.} 

Since the training and test sets of the HM3D dataset \cite{ramakrishnan2021habitat} used by Splatter-360 \cite{chen2024splatter} are not publicly available, we follow their dataset generation process using AI-Habitat \cite{savva2019habitat} to construct our own HM3D training and test sets. Specifically, we render videos along random camera trajectories and generate panoramic images by stitching six cube maps for each viewpoint. For other datasets, we directly use the available off-the-shelf data.

\section{Additional Quantitative Comparisons}

We conducted additional experiments to assess and compare the models' extrapolation capabilities. Following the testing procedure outlined in the main text, we fixed the input frame interval at 100 and randomly selected test frames from a 50-frame range beyond the two input frames. The results, presented in Table \ref{tab:extrapolation}, show that despite all methods being trained with supervision on frames between the inputs, our model consistently outperforms others across all metrics—except for SSIM on the HM3D dataset. This strong performance across most metrics indicates that our model learns more robust spatial representations, enabling more accurate extrapolation beyond the training distribution.

\begin{table}[t]
\centering
\renewcommand{\arraystretch}{1.0}
\setlength{\tabcolsep}{4pt}
\caption{\textbf{Quantitative comparison in view extrapolation.} Methods are evaluated on the Replica and HM3D datasets.}
\vspace{-10pt}
\resizebox{\columnwidth}{!}{
\begin{tabular}{lccccc}
    \toprule
    \textbf{Dataset} & \textbf{Metric} & \textbf{MVSplat} & \textbf{Splatter-360} & \textbf{PanoSplatt3R} \\
    \midrule
    \multirow{6}{*}{\textbf{Replica}~\cite{straub2019replica}} &
    PSNR$\uparrow$ & 27.188 & 26.975 & \textbf{29.371}\\        
    & SSIM$\uparrow$ & 0.895 & 0.904 & \textbf{0.914} \\
    & LPIPS$\downarrow$ & 0.135 & 0.123 & \textbf{0.107} \\ 
    & Abs Rel$\downarrow$ & 0.130 & 0.095 & \textbf{0.059} \\
    & RMSE$\downarrow$ & 0.313 & 0.277 & \textbf{0.176} \\
    & $\delta < 1.25$$\uparrow$ & 85.007 & 90.648 & \textbf{95.414} \\
    \midrule
    \multirow{6}{*}{\textbf{HM3D}~\cite{ramakrishnan2021habitat}} 
    & PSNR$\uparrow$ & 25.728 & 24.986 & \textbf{26.306}\\
    & SSIM$\uparrow$ & 0.827 & \textbf{0.831} & 0.822\\
    & LPIPS$\downarrow$ & 0.205 & 0.193 & \textbf{0.191}\\     
    & Abs Rel$\downarrow$ & 0.129 & 0.136 & \textbf{0.098} \\
    & RMSE$\downarrow$ & 0.339 & 0.334 & \textbf{0.229} \\
    & $\delta < 1.25$$\uparrow$ & 85.390 & 86.258 & \textbf{92.526} \\
    \bottomrule
\end{tabular}
} 
\label{tab:extrapolation}
\vspace{-15pt}
\end{table}
\section{Visualization Results}

\begin{figure*}[tp]
\centering
\renewcommand{\arraystretch}{1}
\setlength{\tabcolsep}{2pt}
\resizebox{\textwidth}{!}{
\begin{tabular}{ccccc}


\rowname{View 1} &
\resultcellThreeCol{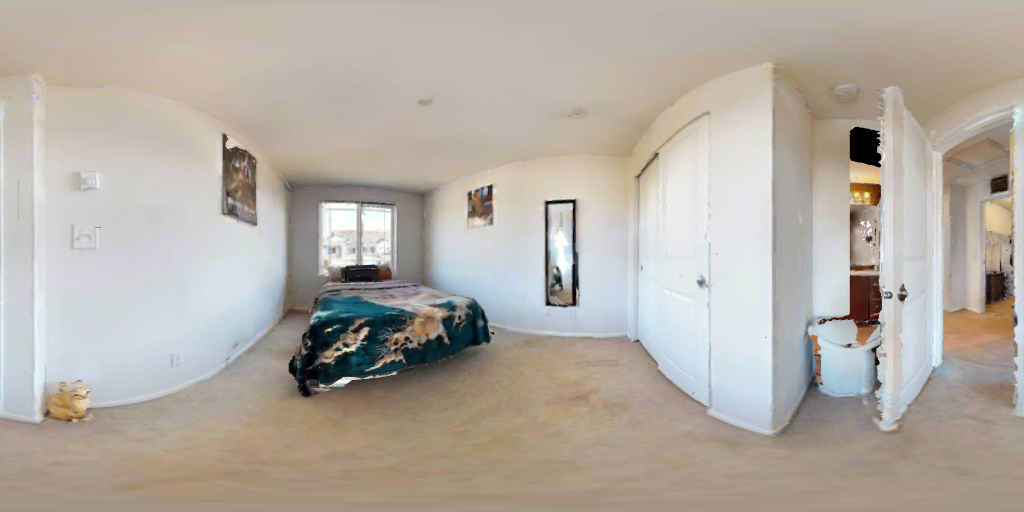} &
\resultcellThreeCol{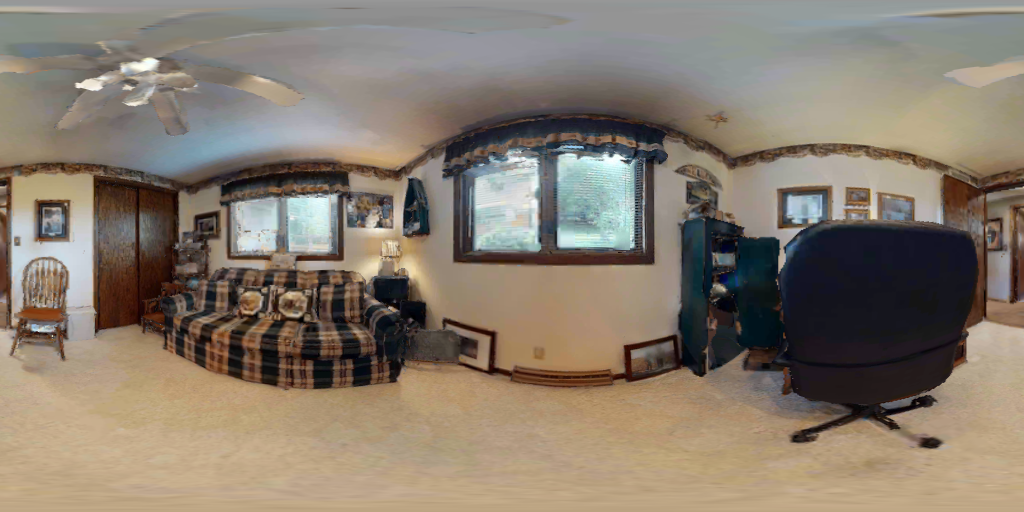} & 
\resultcellThreeCol{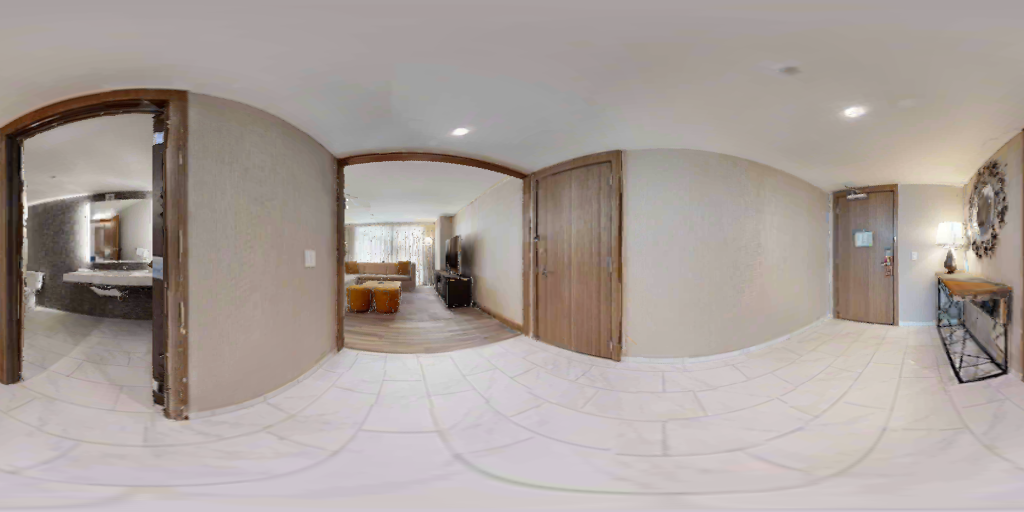} \\

\rowname{View 2} &
\resultcellThreeCol{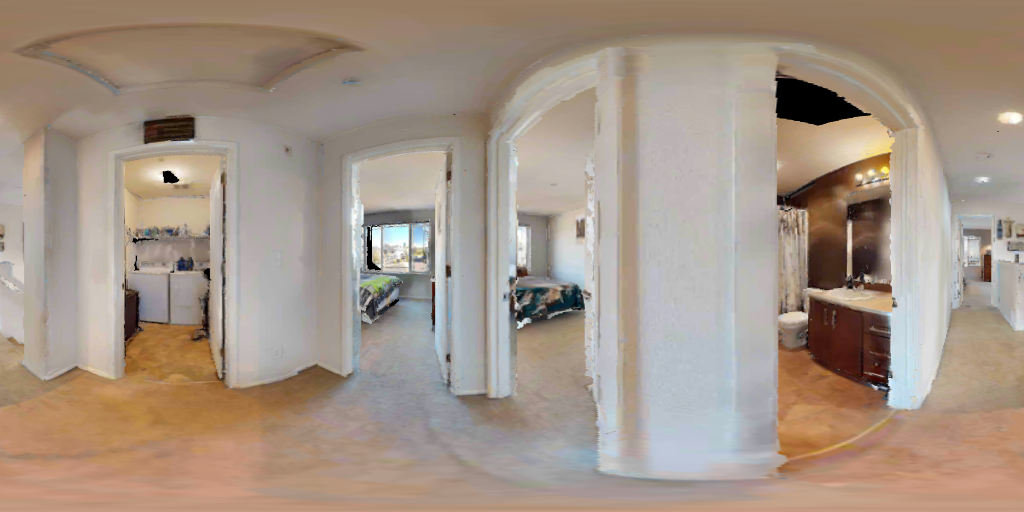} &
\resultcellThreeCol{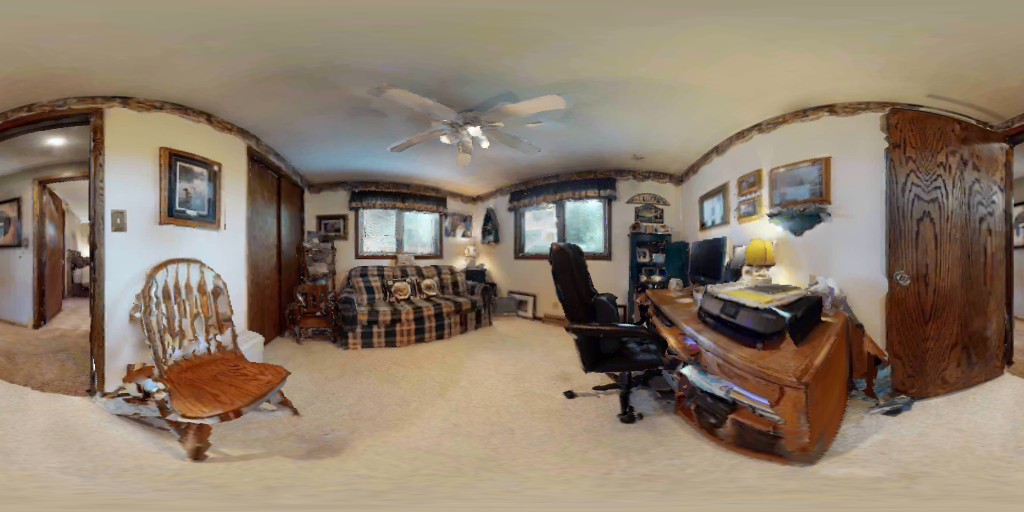} & 
\resultcellThreeCol{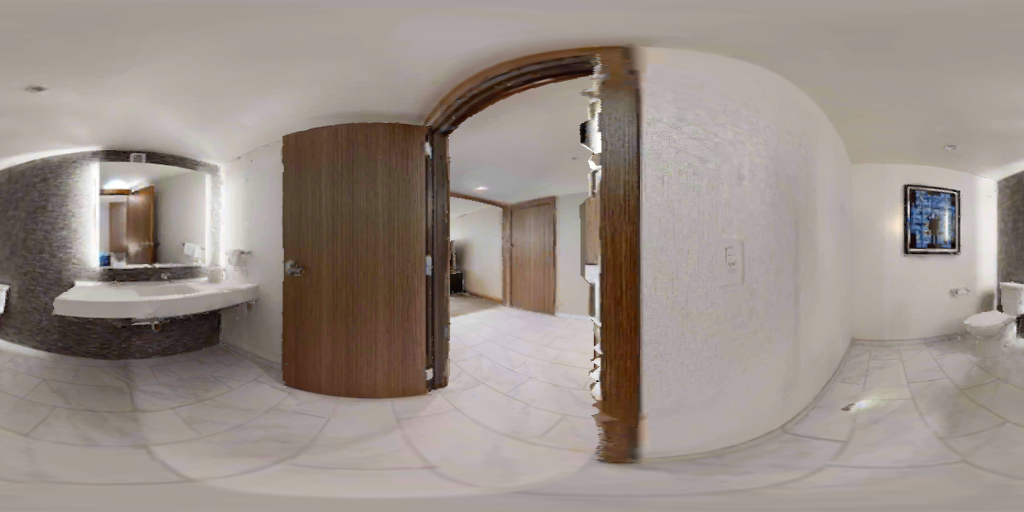} \\ \hline \\ [-8pt]

\rowname{PanoGRF\cite{chen2023panogrf}}  &
\resultcellThreeCol{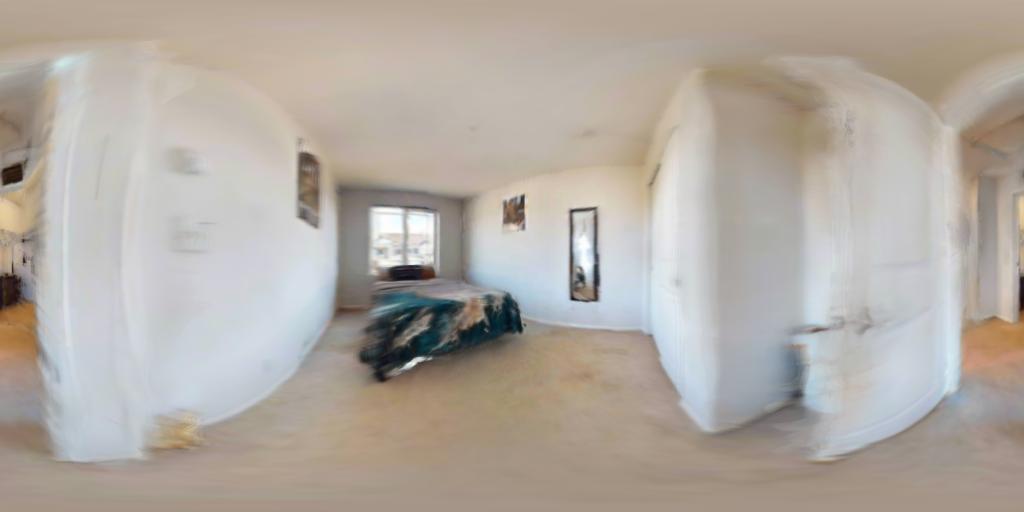} &
\resultcellThreeCol{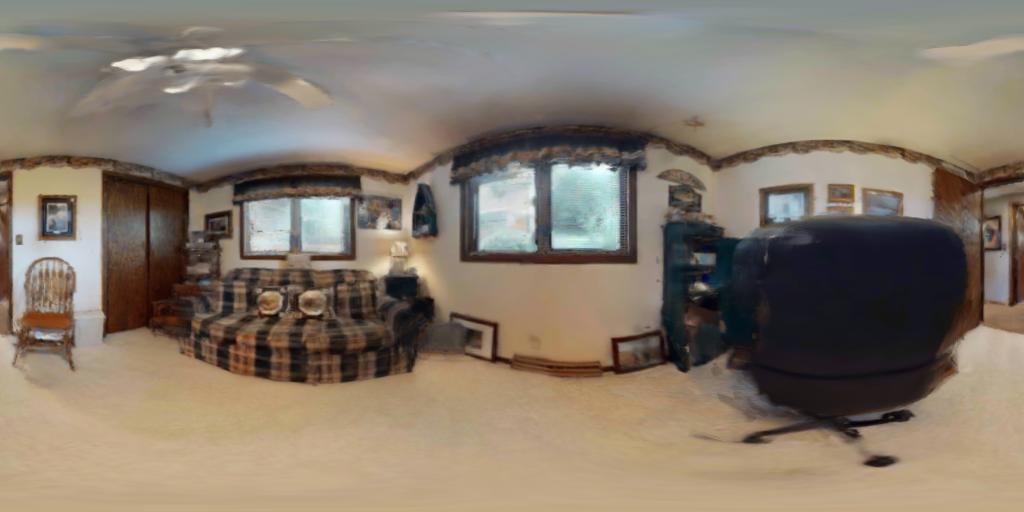} &
\resultcellThreeCol{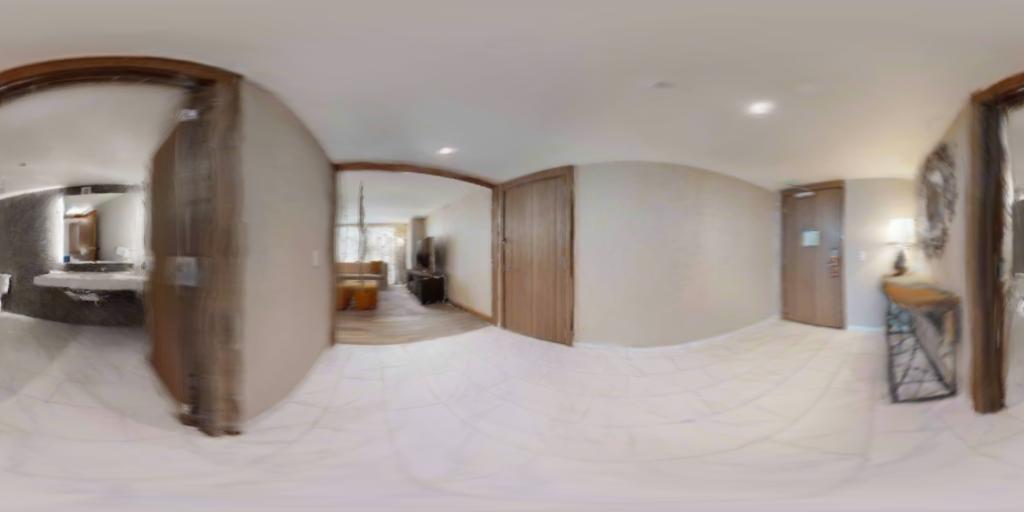} \\

\rowname{MVSplat\cite{mvsplat}}  &
\resultcellThreeCol{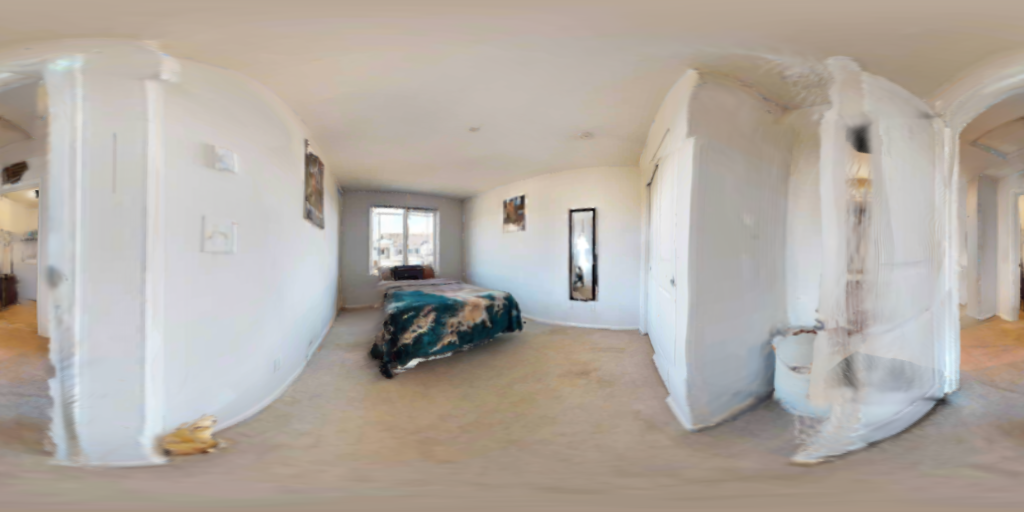} &
\resultcellThreeCol{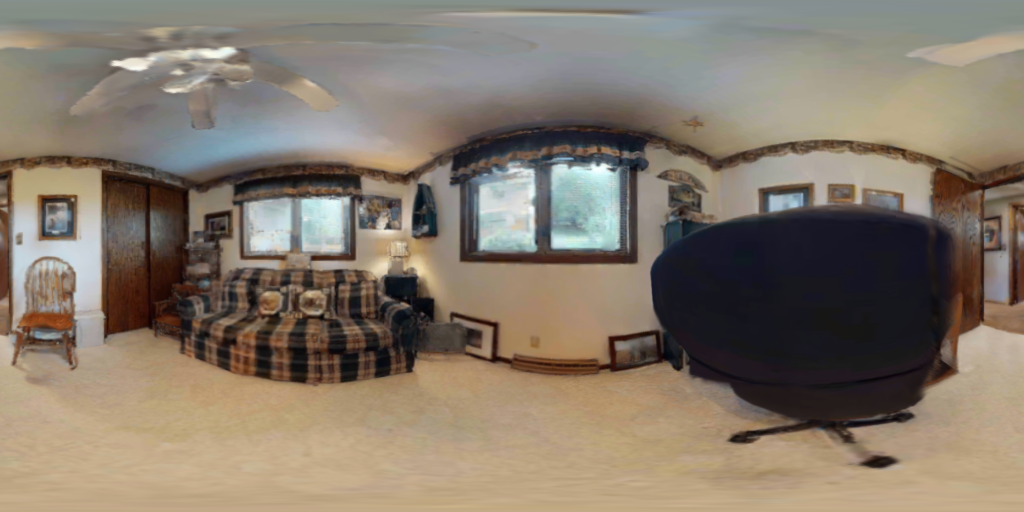} &
\resultcellThreeCol{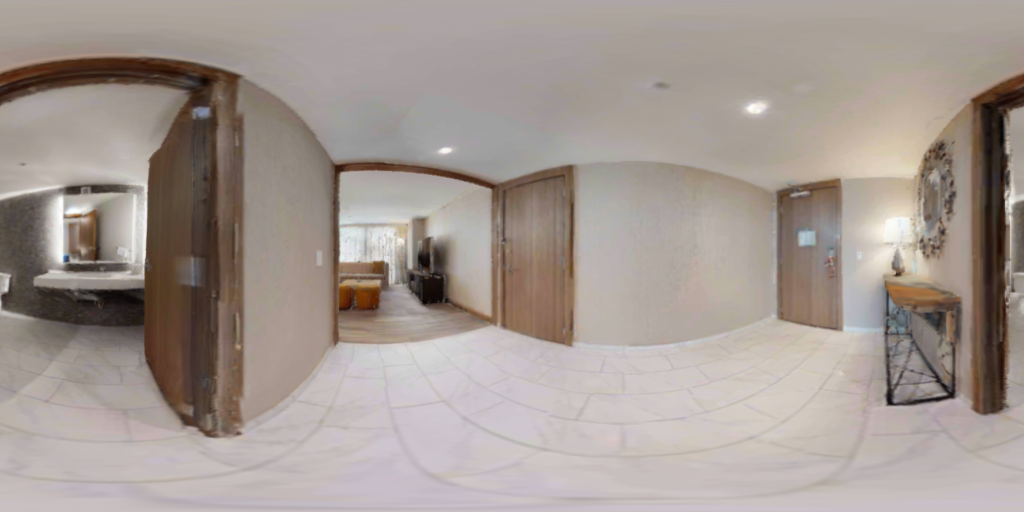} \\

\rowname{Splatter-360\cite{chen2024splatter}} &
\resultcellThreeCol{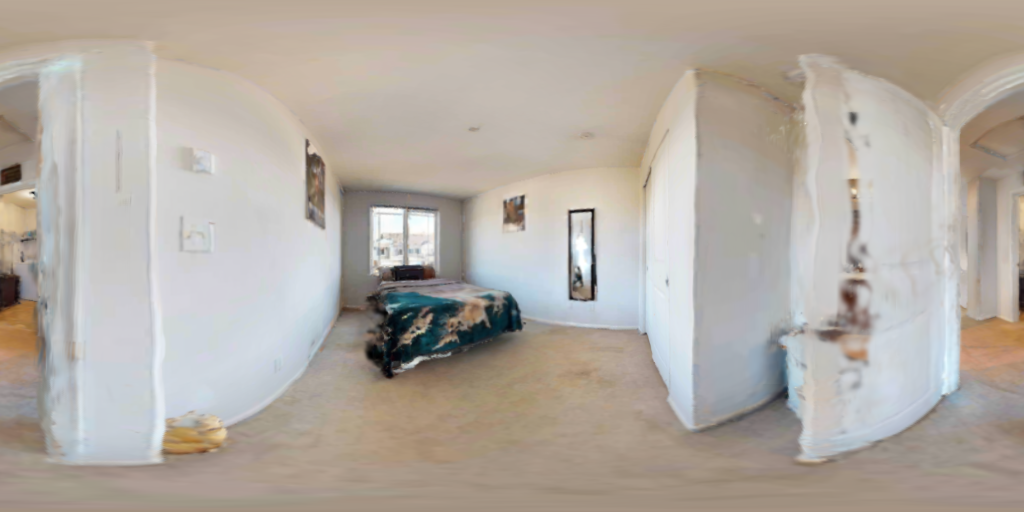} &
\resultcellThreeCol{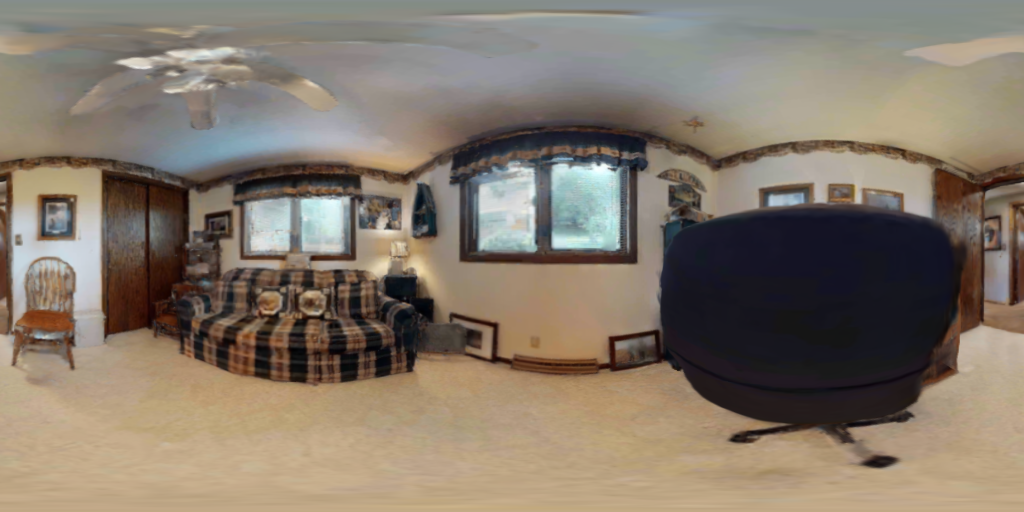} &
\resultcellThreeCol{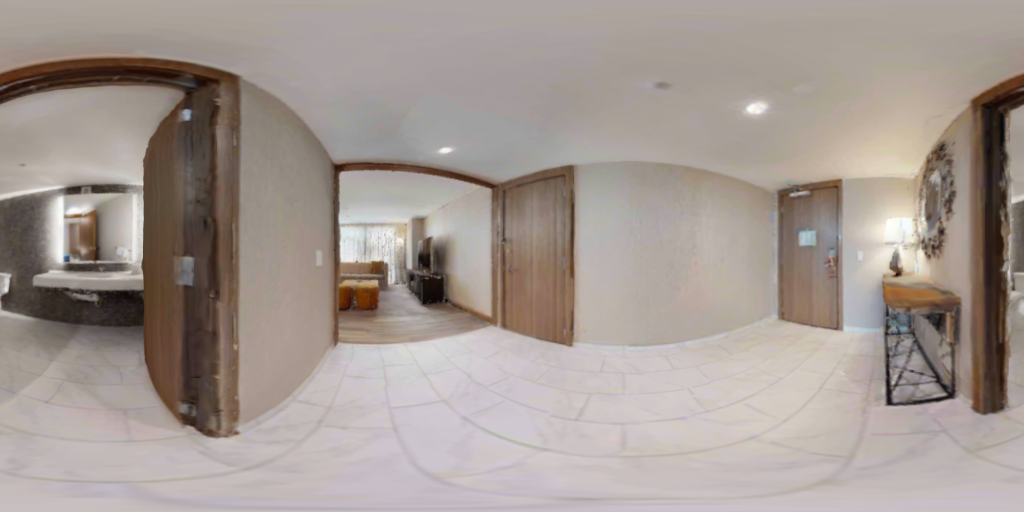} \\

\rowname{PanoSplatt3R} &
\resultcellThreeCol{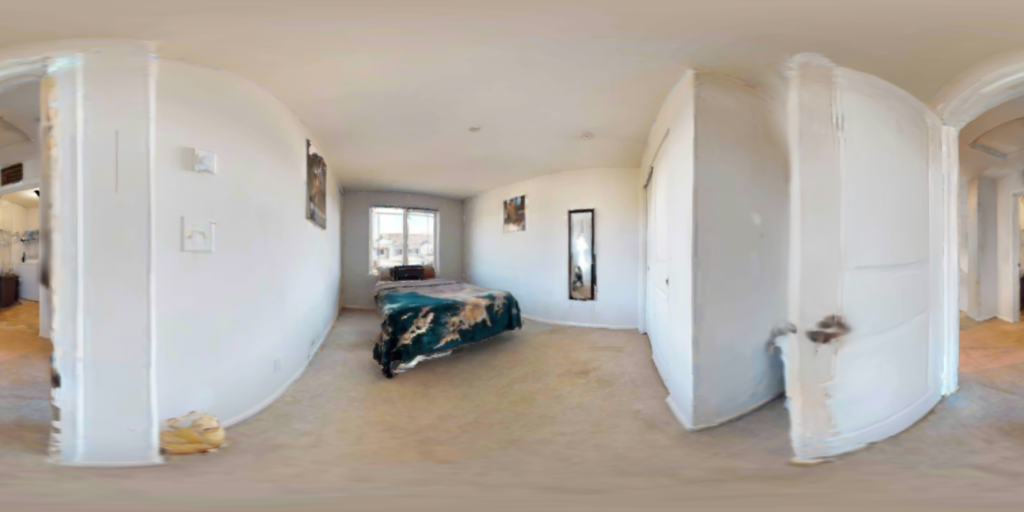} &
\resultcellThreeCol{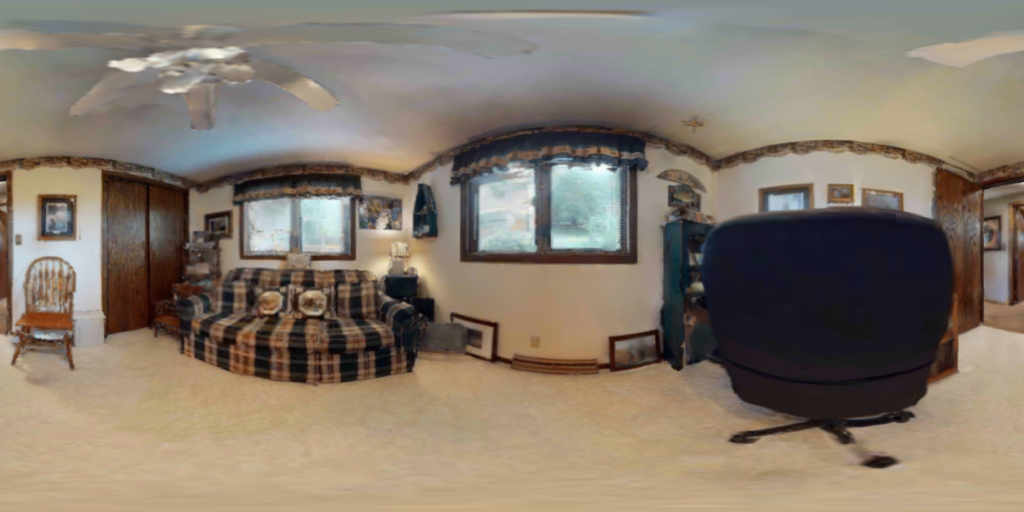} &
\resultcellThreeCol{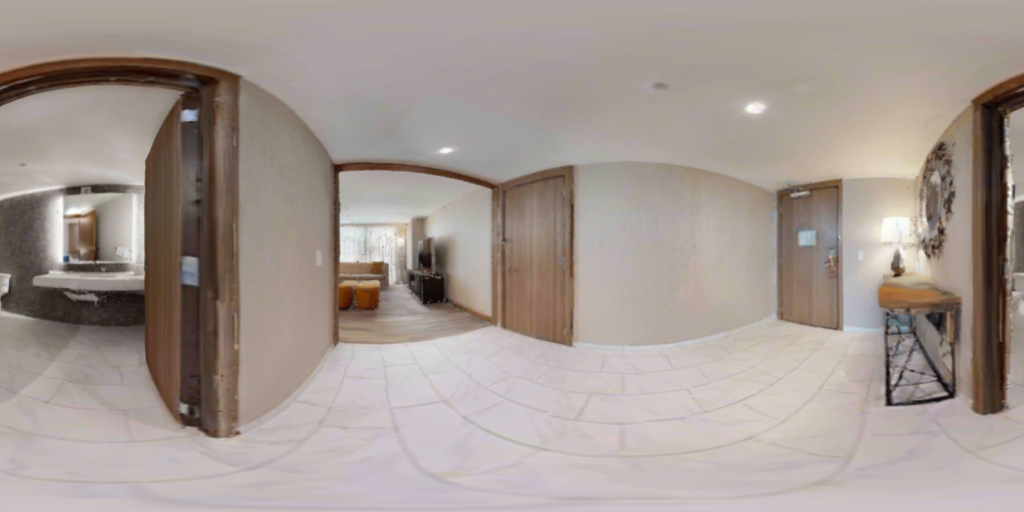} \\ 

&
\hspace{100pt}$\updownarrow$ &
\hspace{60pt}$\updownarrow$ &
\hspace{-100pt}$\updownarrow$ \\ [-12pt]

&
\multicolumn{1}{c}{\textit{unbroken door structure}~~~~~} &
\multicolumn{1}{c}{\textit{correct chair shape}~~~~~~~~~~~~} &
\multicolumn{1}{c}{\textit{accurate hinge detail}} \\ [4pt]

\rowname{GT} &
\resultcellThreeCol{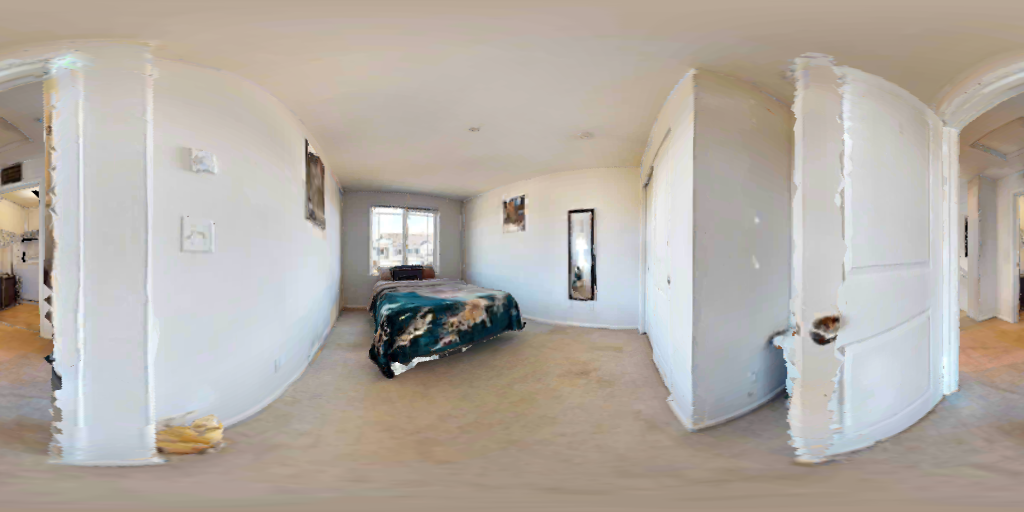} &
\resultcellThreeCol{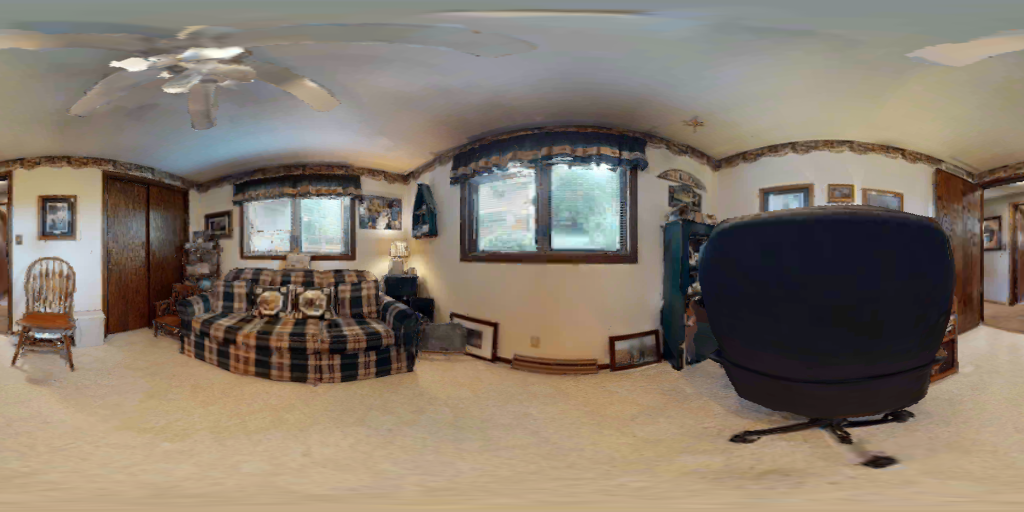} &
\resultcellThreeCol{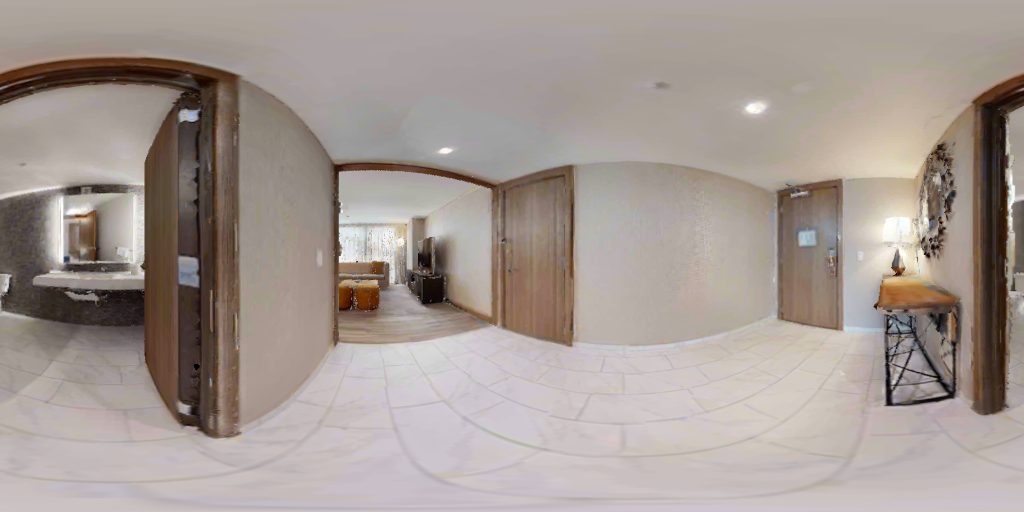} \\

\end{tabular}
} 
\caption{Visual comparisons between PanoGRF, MVSplat, Splatter-360 and PanoSplatt3R(ours) on the HM3D dataset.}
\label{fig:visual_hm3d}
\end{figure*}

\begin{figure*}[tp]
\centering
\renewcommand{\arraystretch}{1}
\setlength{\tabcolsep}{2pt}
\resizebox{\textwidth}{!}{
\begin{tabular}{cccc}

\rowname{View 1} &
\resultcellThreeCol{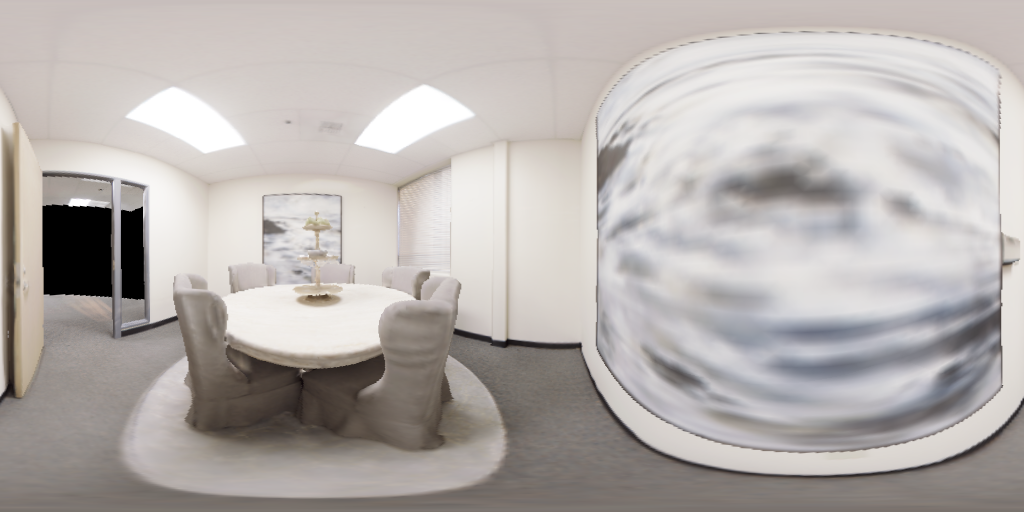} &
\resultcellThreeCol{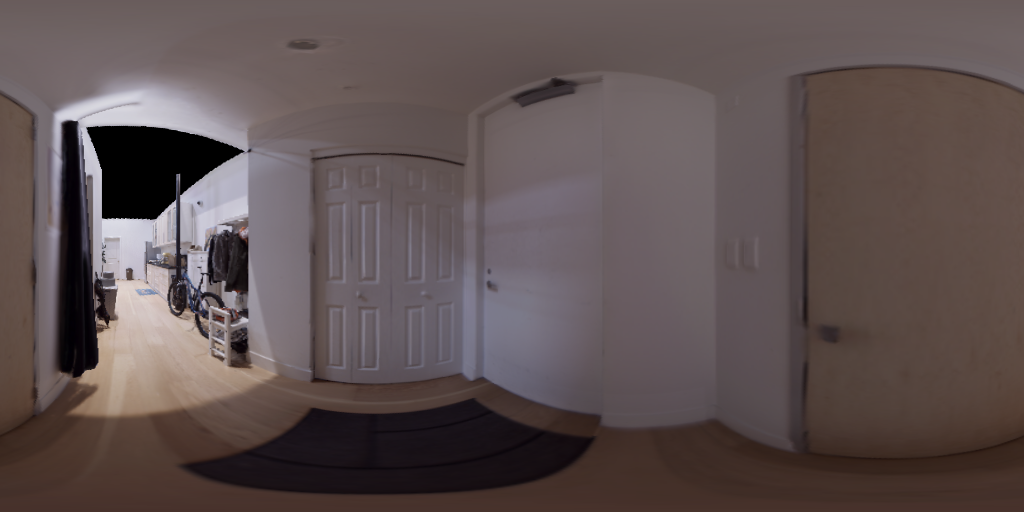} &
\resultcellThreeCol{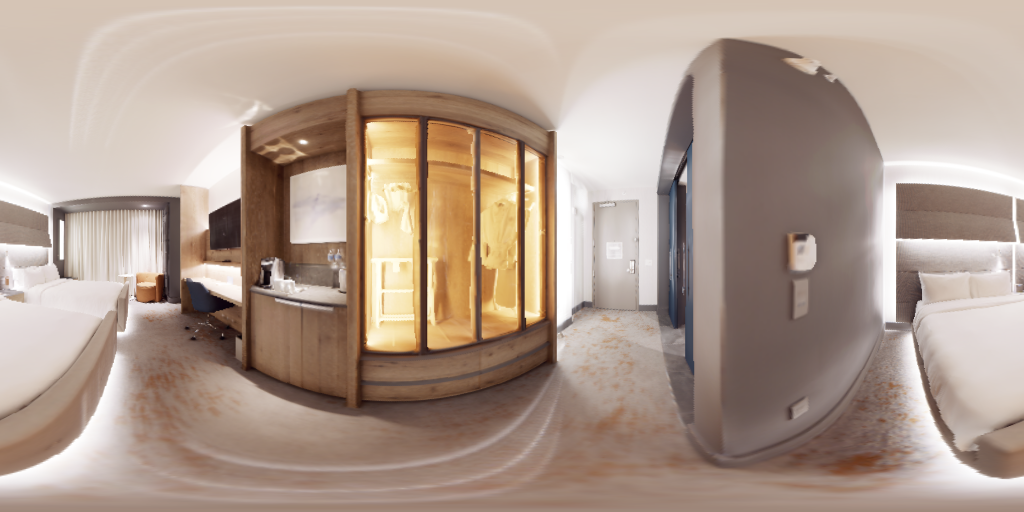} \\

\rowname{View 2} &
\resultcellThreeCol{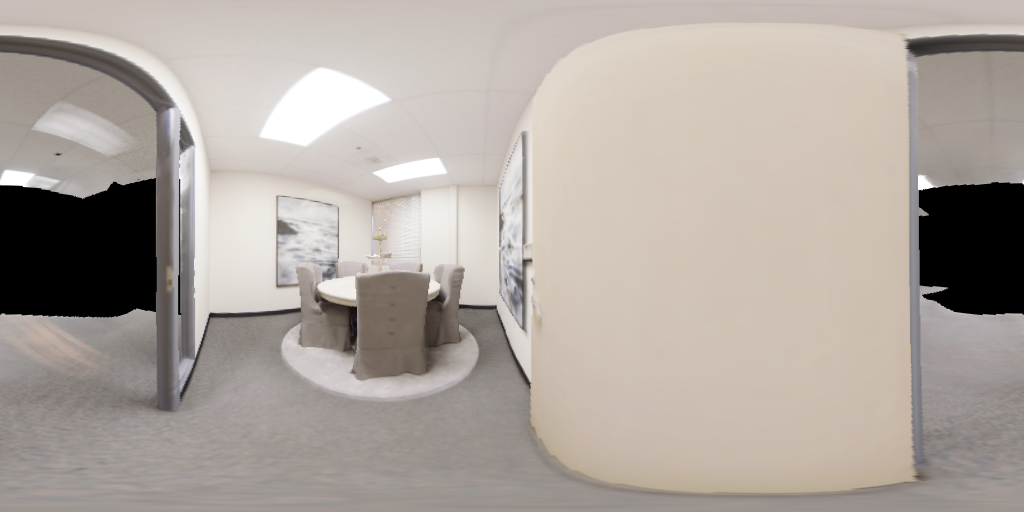} &
\resultcellThreeCol{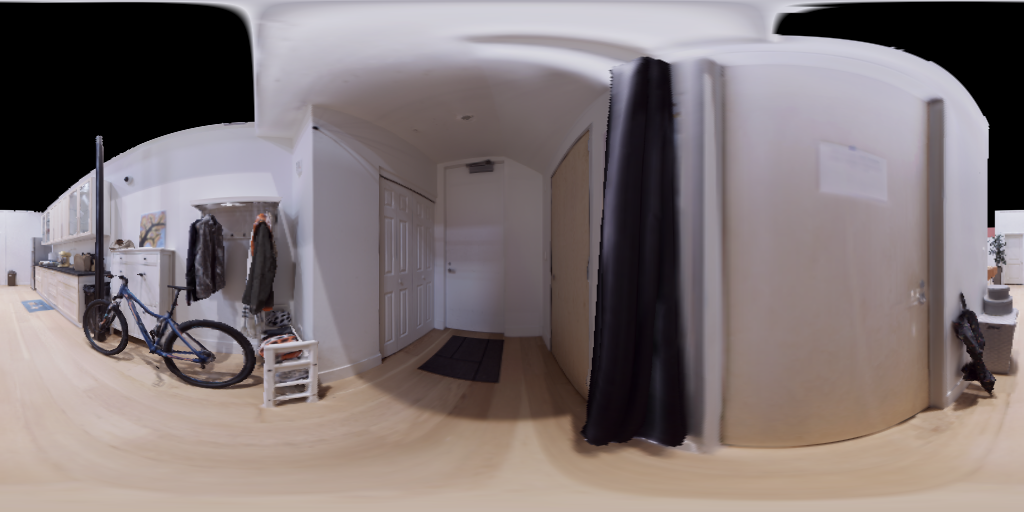} &
\resultcellThreeCol{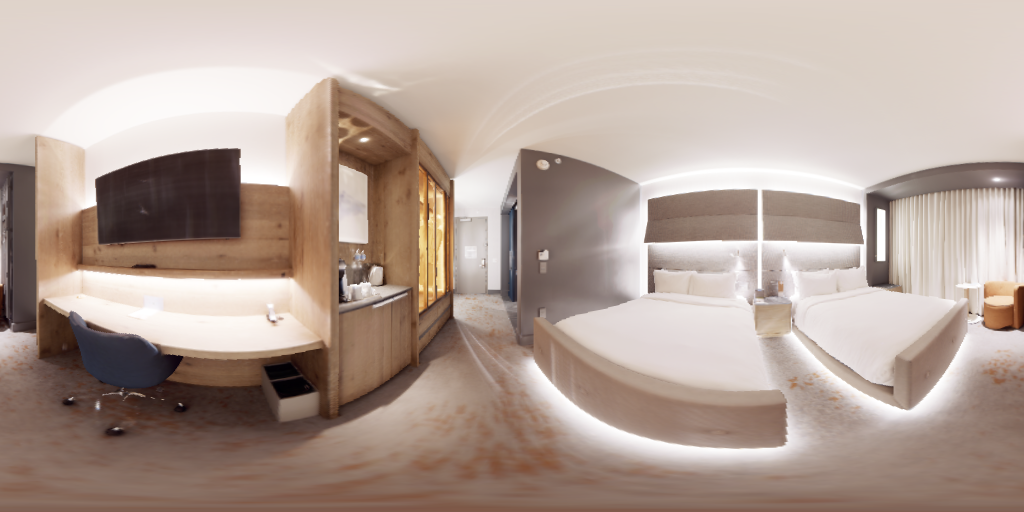} \\ \hline \\[-8pt]

\rowname{PanoGRF\cite{chen2023panogrf}} &
\resultcellThreeCol{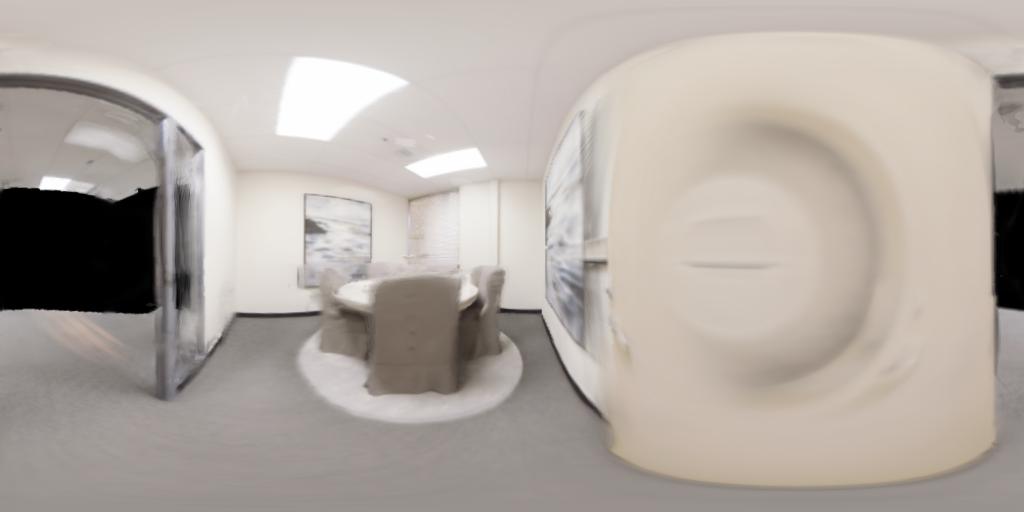} &
\resultcellThreeCol{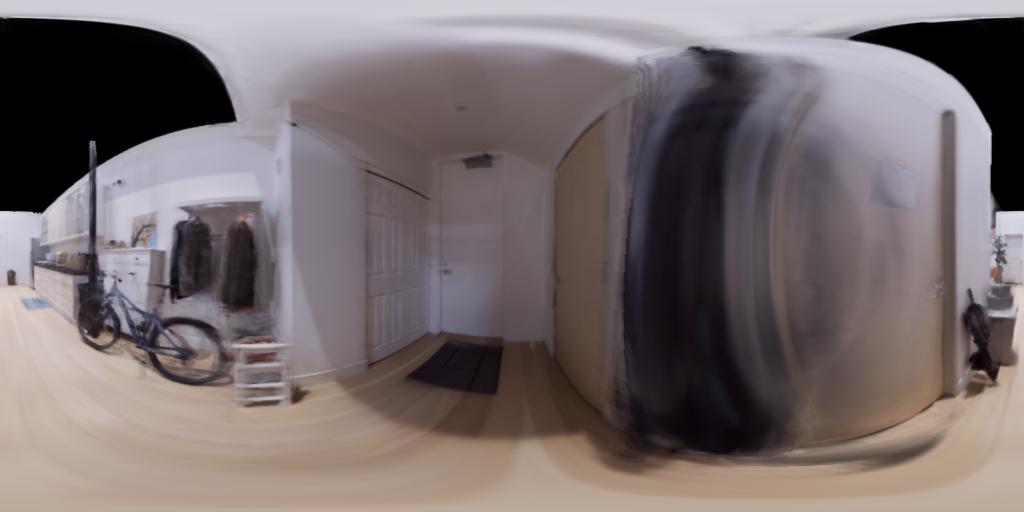} &
\resultcellThreeCol{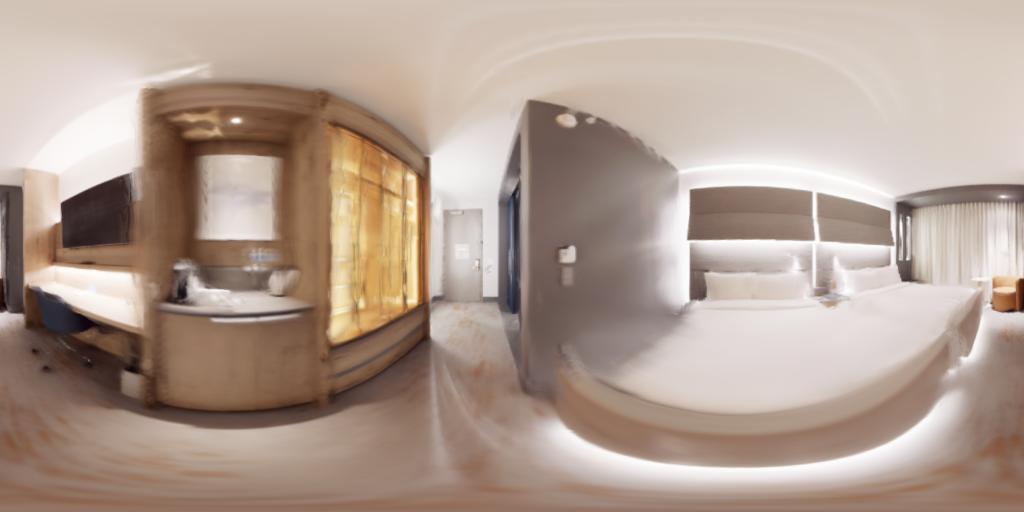} \\

\rowname{MVSplat\cite{mvsplat}} &
\resultcellThreeCol{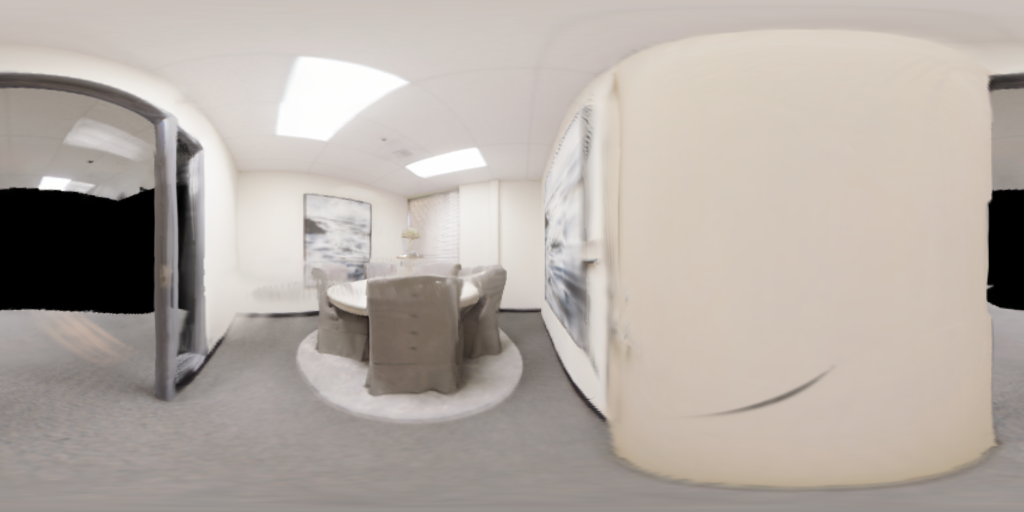} &
\resultcellThreeCol{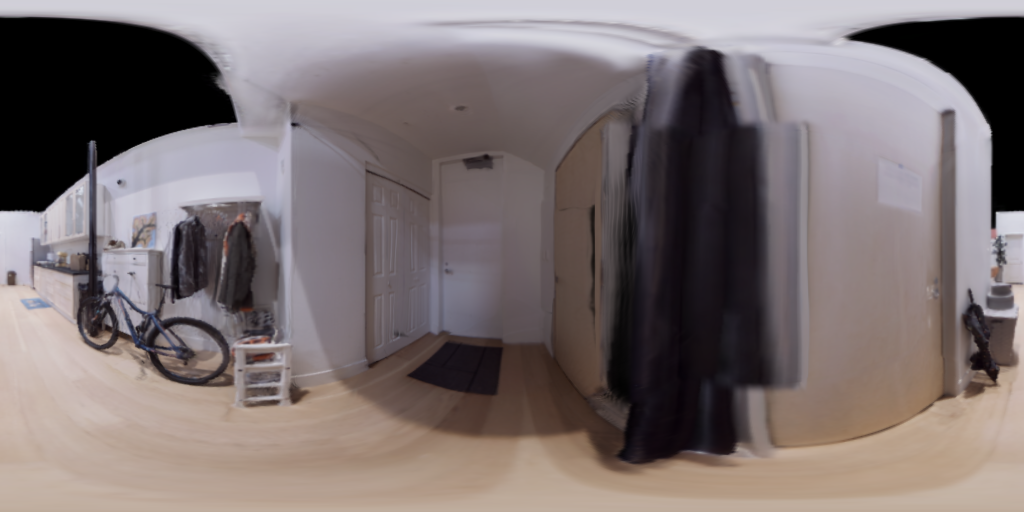} &
\resultcellThreeCol{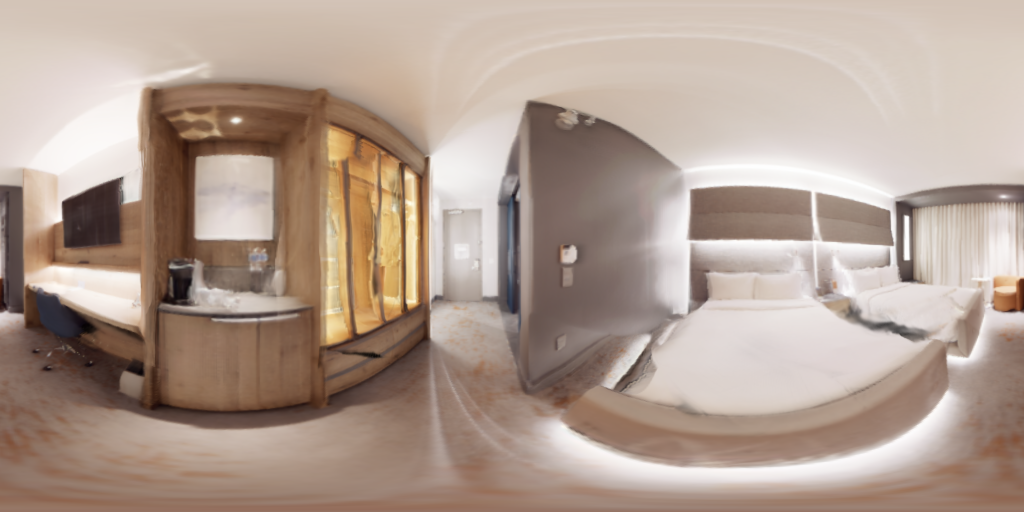} \\

\rowname{Splatter-360\cite{chen2024splatter}} &
\resultcellThreeCol{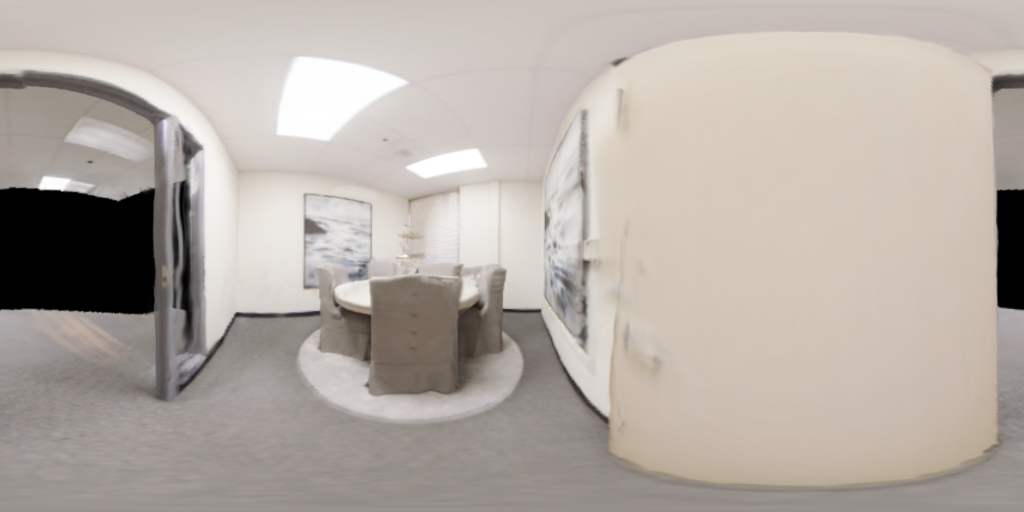} &
\resultcellThreeCol{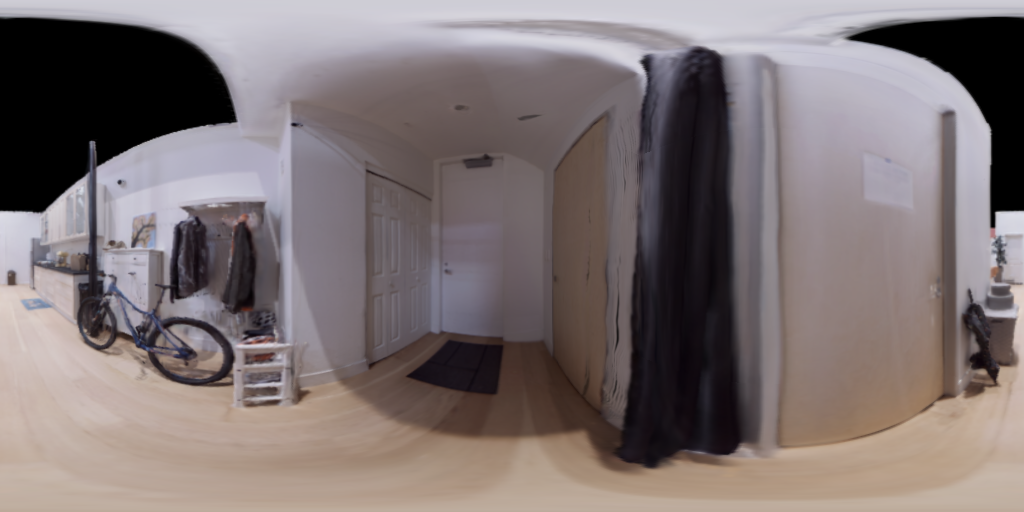} &
\resultcellThreeCol{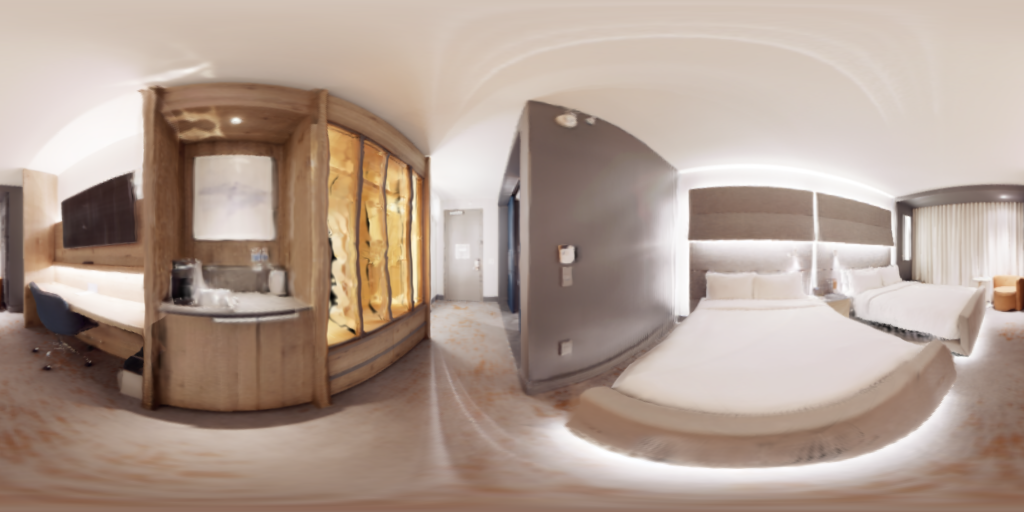} \\

\rowname{PanoSplatt3R} &
\resultcellThreeCol{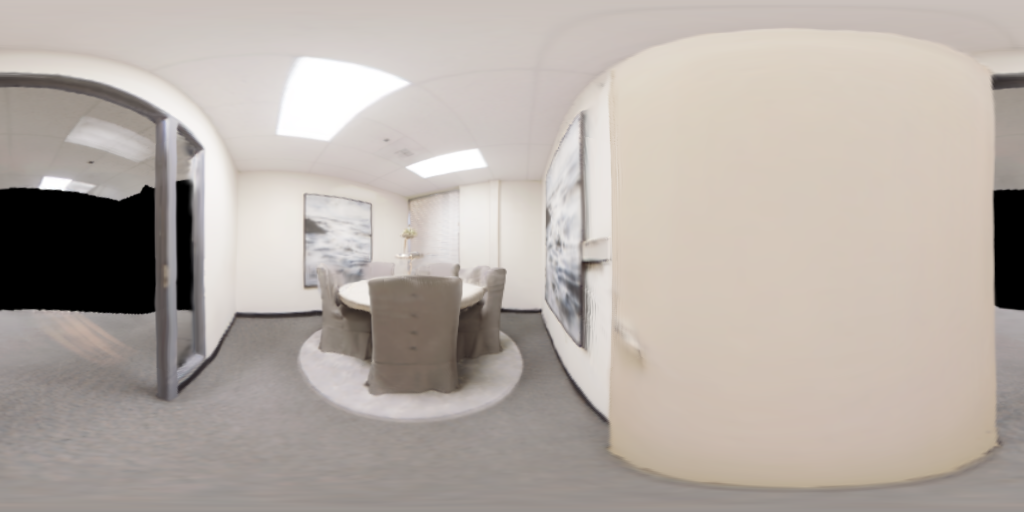} &
\resultcellThreeCol{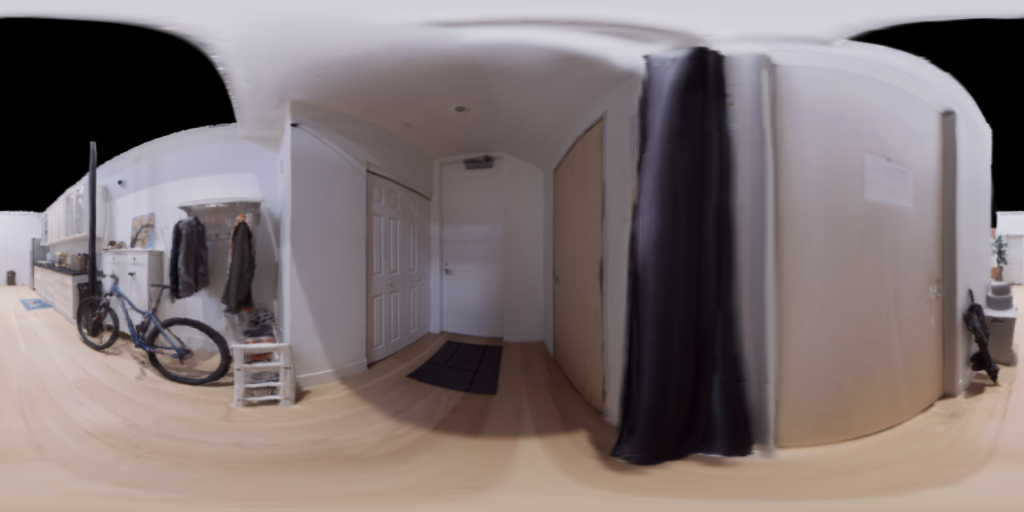} &
\resultcellThreeCol{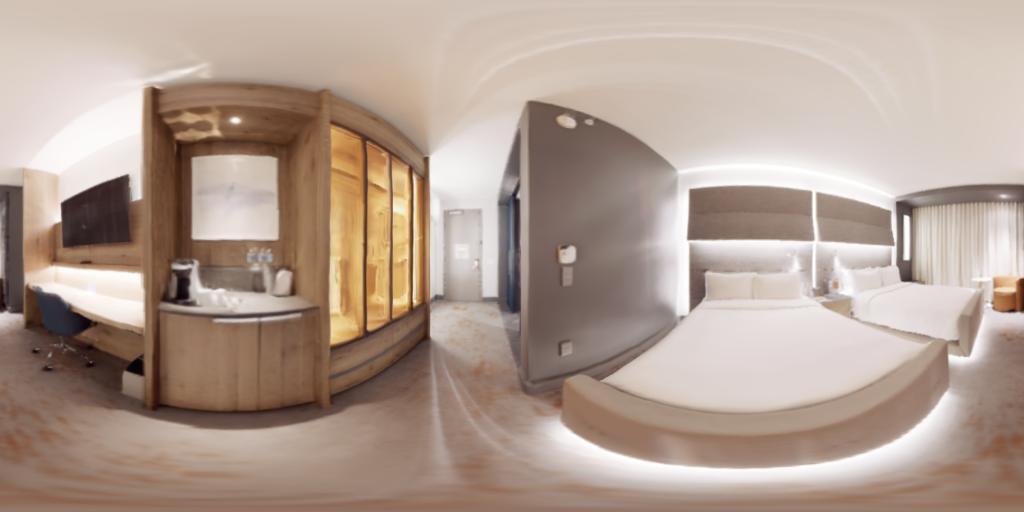} \\

&
\hspace{30pt}$\updownarrow$ &
\hspace{30pt}$\updownarrow$ &
\hspace{40pt}$\updownarrow$ \hspace{70pt} $\updownarrow$ \\ [-12pt]

&
\multicolumn{1}{l}{~~~~~~~~\textit{sharp boundaries ~~ clean wall}} &
\multicolumn{1}{l}{~~~~~~\textit{complete structure}} &
\multicolumn{1}{c}{~~~~~~~~~~~~~~~~\textit{no crooked lines}} \\ [4pt]

\rowname{GT} &
\resultcellThreeCol{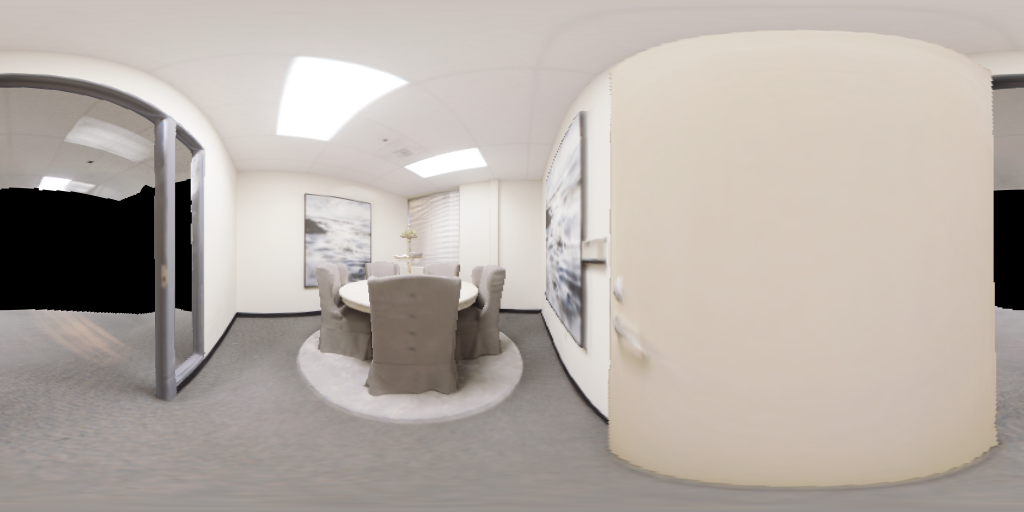} &
\resultcellThreeCol{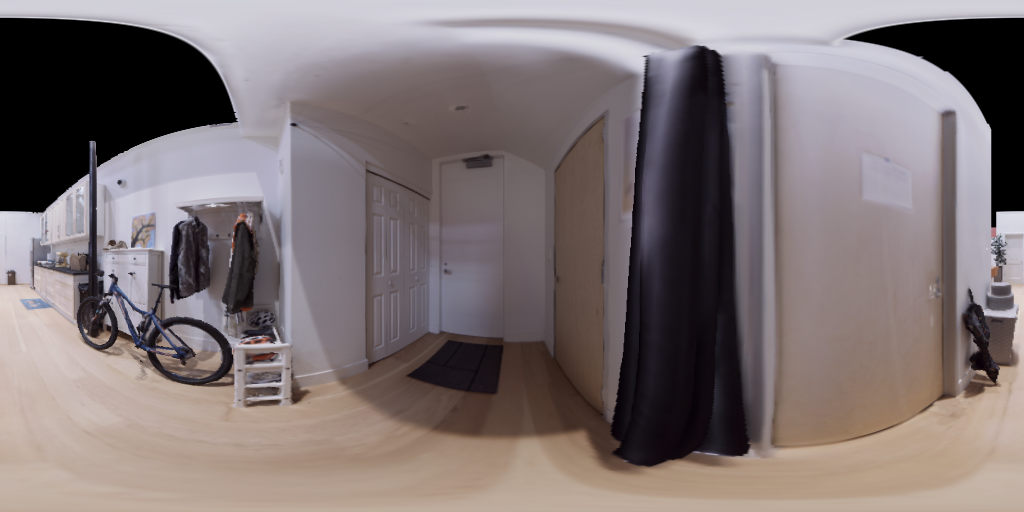} &
\resultcellThreeCol{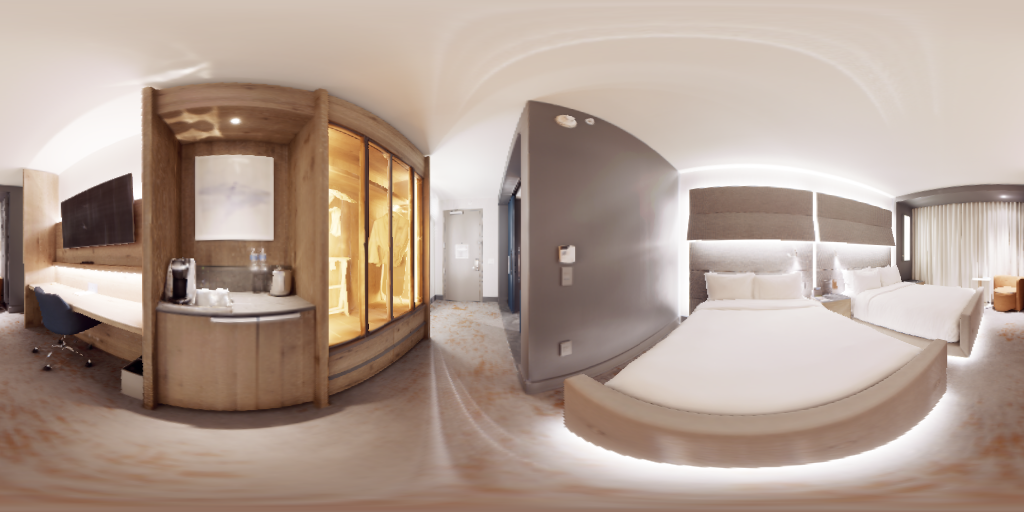} \\

\end{tabular}
} 
\caption{Visual comparisons between PanoGRF, MVSplat, Splatter-360 and PanoSplatt3R(ours) on the Replica dataset.}
\label{fig:visual_replica}
\end{figure*}

We provide visual comparisons of synthesized panoramic images on the HM3D \cite{ramakrishnan2021habitat} and Replica \cite{straub2019replica} datasets, showcasing the performance of different methods. As shown in Figure~\ref{fig:visual_hm3d} and Figure~\ref{fig:visual_replica}, PanoSplatt3R produces the most visually consistent and realistic results, with sharper details, fewer artifacts, and improved structural coherence compared to existing methods. 

Notably, our approach better preserves fine textures and geometric continuity, especially in challenging regions such as object boundaries and occlusions. Please zoom in to examine the details in complex areas, such as edges and textures, where our method demonstrates clear advantages over the others.


\end{document}